\documentclass{article}
\usepackage{fullpage}
\usepackage[utf8]{inputenc} 
\usepackage[T1]{fontenc}    
\usepackage{url}            
\usepackage{booktabs}       
\usepackage{amsfonts}       
\usepackage{nicefrac}       
\usepackage{microtype}      
\usepackage{soul}
\usepackage{graphicx}
\usepackage{amsmath}
\usepackage{outlines}
\usepackage{xurl}
\usepackage{pifont}
\usepackage{caption}

\usepackage[colorlinks=true,
citecolor=blue,
filecolor=black,
linkcolor=blue,
urlcolor=blue]{hyperref}

\usepackage{amsmath,amsfonts,bm}

\usepackage{xcolor}
\usepackage{colortbl}

\def\eqref#1{equation~\ref{#1}}

\DeclareMathAlphabet{\mathsfit}{\encodingdefault}{\sfdefault}{m}{sl}
\SetMathAlphabet{\mathsfit}{bold}{\encodingdefault}{\sfdefault}{bx}{n}

\usepackage{multirow}
\usepackage{paralist}
\usepackage{blindtext}
\usepackage{hyperref}
\usepackage{multicol}
\usepackage{diagbox}

\usepackage[ruled,noend]{algorithm2e}

\SetCommentSty{mycommfont}

\usepackage{here}

\usepackage{amsmath,amssymb,amsfonts,amsbsy,amsfonts,latexsym}
\usepackage{makecell}
\usepackage{xcolor}
\usepackage{colortbl}

\usepackage{tabularx,colortbl,xcolor}
\usepackage[normalem]{ulem}
\useunder{\uline}{\ul}{}

\usepackage{enumitem}

\usepackage{xparse}

\SetKwInput{KwInput}{Input}
\SetKwInput{KwRequire}{Require}

\usepackage{longtable}

\NewDocumentCommand{\var}{O{s} m O{}}{%
  \ensuremath{#1_{#2}^{#3}}
}
\usepackage{siunitx}

\newcommand{\commentout}[1]{}

\definecolor{light-gray}{gray}{0.80}

\newcommand\appref{Appendix~\ref}

\newcommand\fref{Figure~\ref}
\newcommand\tref{Table~\ref}
\newcommand\sref{\textsection~\ref}

\usepackage{amsthm}

\usepackage{amsthm}

\newtheorem{mytheorem}{Theorem}
\newtheorem{assumption}[mytheorem]{Assumption}
\newtheorem{lemma}[mytheorem]{Lemma}
\newtheorem{remk}[mytheorem]{Remark}

\def\bertbase{BERT$_{\text{base}}$\xspace}

\newcommand{\cmark}{\ding{51}}%
\newcommand{\xmark}{\ding{55}}%
\newcommand{\OURS}{XTC\xspace}

\newcommand{\bert}{BERT\xspace}
\newcommand{\onestage}{1S-KD\xspace}
\newcommand{\twostage}{2S-KD\xspace}
\newcommand{\threestage}{3S-KD\xspace}

\usepackage{wrapfig}
\usepackage{chngcntr}
\usepackage{adjustbox}
\usepackage{arydshln}

\usepackage{tcolorbox}

\begin{document}

\title{Extreme Compression for Pre-trained Transformers\\ Made Simple and Efficient}

\author{
 Xiaoxia Wu\thanks{Equal contribution. Code will be released soon as a part of \url{https://github.com/microsoft/DeepSpeed}}~,  Zhewei Yao$^*$,  Minjia Zhang$^*$\\ Conglong Li, Yuxiong He  
 \vspace{0.2cm} 
 \\
Microsoft \\
\texttt{\{xiaoxiawu, zheweiyao, minjiaz, conglong.li, yuxhe\}}@microsoft.com
}

\date{}
\maketitle

\begin{abstract}
Extreme compression, particularly ultra-low bit precision (binary/ternary) quantization, has been proposed to fit large NLP models on resource-constraint devices. 
However, to preserve the accuracy for such aggressive compression schemes, cutting-edge methods usually introduce complicated compression pipelines, e.g., multi-stage expensive knowledge distillation with extensive hyperparameter tuning. 
Also, they oftentimes focus less on smaller transformer models that have already been heavily compressed via knowledge distillation and lack a systematic study to show the effectiveness of their methods.
In this paper, we perform a very comprehensive systematic study to measure the impact of many key hyperparameters and training strategies from previous works. 
As a result, we find out that previous baselines for ultra-low bit precision quantization are significantly under-trained. 
Based on our study, we propose a simple yet effective compression pipeline for extreme compression, named \OURS. 
\OURS demonstrates that
(1) we can skip the pre-training knowledge distillation to obtain a 5-layer \bert while achieving better performance than previous state-of-the-art methods, e.g., the 6-layer TinyBERT; 
(2) extreme quantization plus layer reduction is able to reduce the model size by 50x, resulting in new state-of-the-art results on GLUE tasks.
\end{abstract}

\section{Introduction}
\label{sec:intro}

Over the past few years, we have witnessed  the model size has grown at an unprecedented speed, from a few hundred million parameters (e.g., BERT~\cite{bert}, RoBERTa~\cite{roberta}, DeBERTA ~\cite{deberta}, T5~\cite{T5},GPT-2~\cite{gpt-2}) to a few hundreds of billions of parameters (e.g., 175B GPT-3~\cite{gpt-3}, 530B MT-NLT~\cite{mt-nlg}), showing outstanding results on a wide range of language processing tasks. 
Despite the remarkable performance in accuracy, there have been huge challenges to deploying these models, especially on resource-constrained edge or embedded devices. Many research efforts have been made to compress these huge transformer models including knowledge distillation~\cite{jiao-etal-2020-tinybert,mini-lm,minilm-v2,ad2}, pruning~\cite{yao2021mlpruning,sanh2020movement,bert-lottery}, and low-rank decomposition~\cite{ladabert}. 
Orthogonally, quantization focuses on replacing the floating-point weights of a pre-trained Transformer network with low-precision representation. 
This makes quantization particularly appealing when compressing models that have already been optimized in terms of network architecture.

Popular quantization methods include post-training quantization~\cite{shomron2021post,nagel2020up,liu2021post}, quantization-aware training (QAT)~\cite{kim2016bitwise,banner2018scalable,mellempudi2017ternary,juefei2017local}, and their variations~\cite{shen2020q,kim2021bert,dong2019hawq}. 
The former directly quantizes trained model weights from floating-point values to low precision values using a scalar quantizer, which is simple but can induce a significant drop in accuracy. 
To address this issue, quantization-aware training directly quantizes a model during training by quantizing all the weights during the forward and using a straight-through estimator (STE)~\cite{bengio2013estimating} to compute the gradients for the quantizers.

Recently, several QAT works further push the limit of BERT quantization to the extreme via ternarized (2-bit) (\cite{zhang-etal-2020-ternarybert}) and binarized (1-bit) weights (\cite{bai-etal-2021-binarybert}) together with 4-/8-bit quantized activation. 
These compression methods have been referred to as \textbf{extreme quantization} since the limit of weight quantization, in theory, can bring over an order of magnitude compression rates (e.g., 16-32 times). 
One particular challenge identified in \cite{bai-etal-2021-binarybert} was that it was highly difficult to perform binarization as there exits a sharp performance drop from ternarized to binarized networks. 
To address this issue, prior work proposed multiple optimizations where one first trains a ternarized DynaBert~\cite{hou2020dynabert} and then binarizes with weight splitting.
In both phases, multi-stage distillation with multiple learning rates tuning and data augmentation~\cite{jiao-etal-2020-tinybert} are used. 
Prior works claim these optimizations are essential for closing the accuracy gap from binary quantization.

While the above methodology is promising, several unanswered questions are related to these recent extreme quantization methods. 
First, as multiple ad-hoc optimizations are applied at different stages, the compression pipeline becomes very complex and expensive, limiting the applicability of extreme quantization in practice. 
Moreover, a systematical evaluation and comparison of these optimizations are missing, and the underlying question remains open for extreme quantization: 
\begin{center}
\emph{what are the necessities of ad-hoc optimizations to recover the accuracy los?} 
\end{center}

\begin{figure}[t]
    \centering
    \includegraphics[width=1\textwidth]{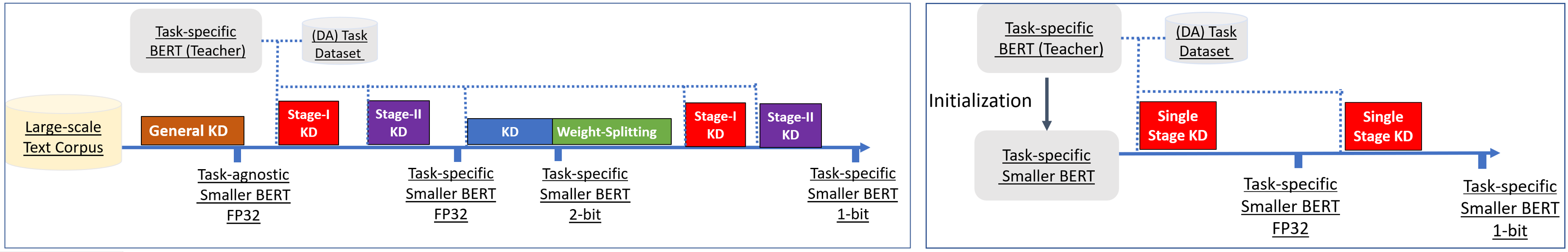}
    \caption{The left figure summarizes how to do 1-bit quantization for a layer-reduced model based on~\cite{jiao-etal-2020-tinybert,bai-etal-2021-binarybert}. 
    It involves expensive pretraining on an fp-32 small model, task-specific training on 32-bit and 2-bit models, weight-splitting, and the final 1-bit model training. 
    Along the way, it applies multi-stage knowledge distillation with data augumentation, which needs considerable hyperparameter tuning efforts. 
    The right figure  is our proposed method, \OURS (see details in \sref{sec:design}), a simple while effective pipeline (see \fref{fig:sota} for highlighted results). 
    Better read with a computer screen.}
    \label{fig:propose-method}
\end{figure}

\begin{wrapfigure}{r}{8cm}
\includegraphics[width=0.5\textwidth]{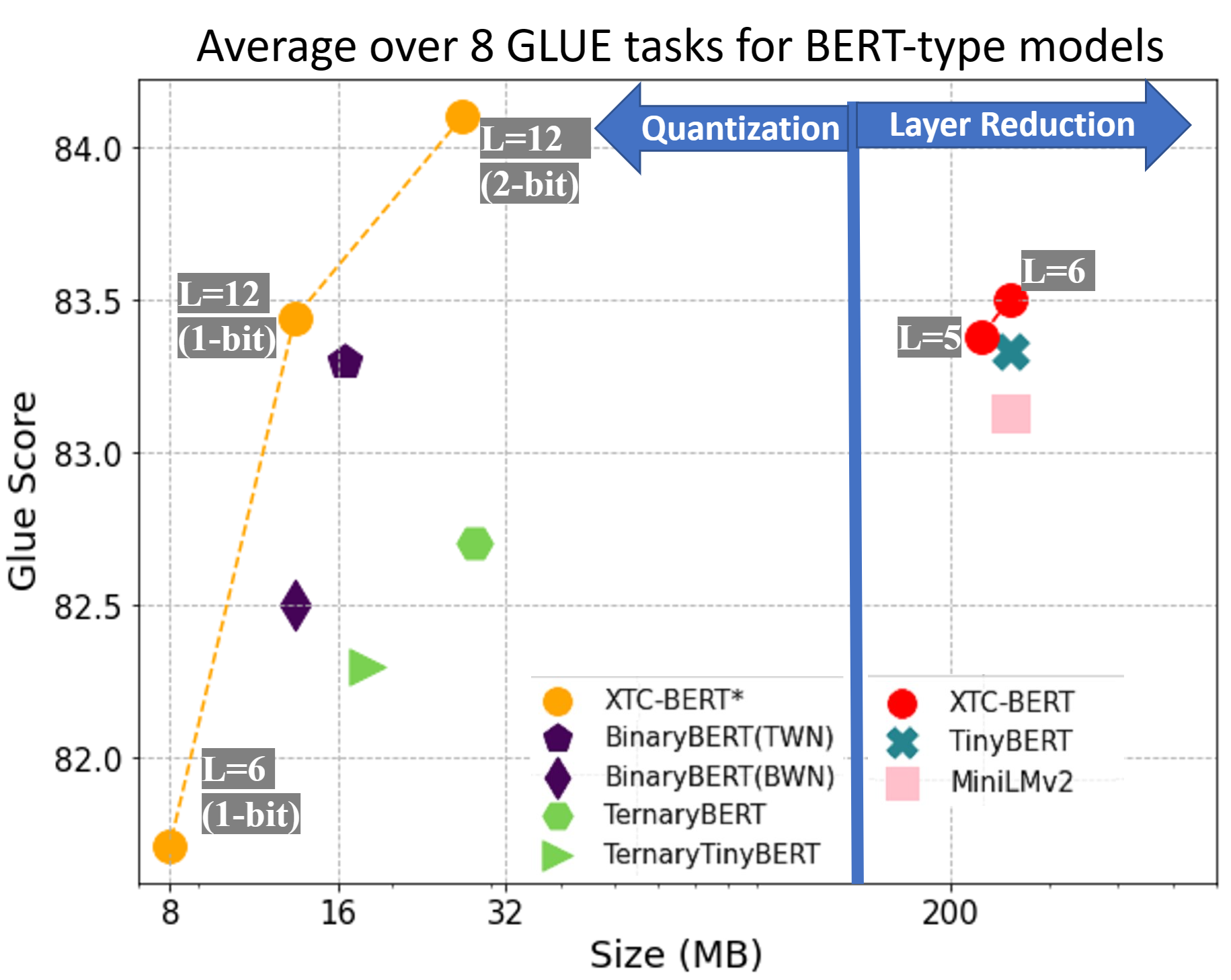}
\caption{The comparison between \OURS with other SOTA results.}
\label{fig:sota}
\end{wrapfigure}

Second, prior extreme quantization primarily focused on reducing the precision of the network. 
Meanwhile, several advancements have also been made in the research direction of knowledge distillation, where large teacher models are used to guide the learning of a small student model. 
Examples include DistilBERT~\cite{distill-bert},  MiniLM~\cite{mini-lm,minilm-v2}, MobileBERT~\cite{sun2020mobilebert}, which demonstrate 2-4$\times$ model size reduction by reducing the depth or width without much accuracy loss through pre-training distillation with an optional fine-tuning distillation. 
Notably,  TinyBERT~\cite{jiao-etal-2020-tinybert} proposes to perform deep distillation in both the pre-training and fine-tuning stages and shows that this strategy achieves state-of-the-art results on GLUE tasks. 
However, most of these were done without quantization, and there are few studies about the interplay of extreme quantization with these heavily distilled models, which poses questions on

\begin{center}
\emph{to what extent, a smaller distilled model benefits from extreme quantization?} 
\end{center}

\paragraph{Contribution.} To investigate the above questions, we make the following contributions:\\ 
(1) We present a systematic study of extreme quantization methods by fine-tuning $\geq$1000 pre-trained Transformer models, which includes a careful evaluation of the effects of hyperparameters and several methods introduced in extreme quantization. \\
(2) We find that previous extreme quantization studies overlooked certain design choices, which lead to under-trained binarized networks and unnecessarily complex optimizations. 
Instead, we derive a celebrating recipe for extreme quantization, which is not only simpler but also allows us to achieve an even larger compression ratio and higher accuracy than existing methods (see~\fref{fig:propose-method}, right). \\
(3) We find that extreme quantization can be effectively combined with lightweight layer reduction, which allows us to achieve greater compression rates for pre-trained Transformers with better accuracy than prior methods while enjoying the additional benefits of flexibly adjusting the size of the student model for each use-case individually, without the expensive pre-training distillation.

Our evaluation results, as illustrated in~\fref{fig:sota}, show that our simple yet effective method can:\\
(1) compress \bertbase to a 5-layer \bertbase while achieving better performance than previous state-of-the-art distillation methods, e.g., the 6-layer TinyBERT~\cite{jiao-etal-2020-tinybert}, without incurring the computationally expensive pre-training distillation; \\
(2) reduce the model size by 50$\times$ via performing robust extreme quantization on lightweight layer reduced models while obtaining better accuracy than prior extreme quantization methods, e.g., the 12-layer 1-bit \bertbase, resulting in new state-of-the-art results on GLUE tasks.

\section{Related Work}
\label{sec:related_work}

Quantization becomes practically appealing as it not only reduces memory bandwidth consumption but also brings the additional benefit of further accelerating inference on supporting hardware~\cite{shen2020q,kim2021bert,dong2019hawq}. 
Most quantization works focus on INT8 quantization~\cite{is-bert-robust} or mixed INT8/INT4 quantization~\cite{shen2020q}. 
Our work differs from those in that we investigate extreme quantization where the weight values are represented with only 1-bit or 2-bit at most. 
There are prior works that show the feasibility of using only ternary or even binary weights~\cite{zhang-etal-2020-ternarybert,bai-etal-2021-binarybert}. 
Unlike those work, which uses multiple optimizations with complex pipelines, our investigation leads us to introduce a simple yet more efficient method for extreme compression with more excellent compression rates.  

On a separate line of research, reducing the number of parameters of deep neural networks models with full precision (no quantization) has been an active research area by applying the powerful knowleadge distillation (KD)~\cite{kd-hinton,mini-lm}, where a stronger teacher model guides the learning of another small student model to minimize the discrepancy between the teacher and student outputs. 
Please see~\appref{sec:KD-review} for a comprehensive literature review on KD. 
Instead of proposing a more advanced distillation method, we perform a well-rounded comparative study on the effectiveness of the recently proposed multiple-stage distillation \cite{jiao-etal-2020-tinybert}.

\section{Extreme Compression Procedure Analysis}
This section presents several studies that have guided the proposed method introduced in Section~\ref{sec:design}. 
All these evaluations are performed with the General Language Understanding Evaluation (GLUE) benchmark~\cite{glue}, which is a collection of datasets for evaluating natural language understanding systems. 
For the subsequent studies, we report results on the development sets after compressing a pre-trained model (e.g., \bertbase and TinyBERT) using the corresponding single-task training data. 

Previous works~\cite{zhang-etal-2020-ternarybert,bai-etal-2021-binarybert} on extreme quantization of transformer models state three hypotheses for what can be related to the difficulty of performing extreme quantization: 
\begin{itemize}\label{subsec:hypothesis-analysis}
    \item Directly training a binarized BERT is complicated due to its irregular loss landscape, so it is better to first to train a ternarized model to initialize the binarized network. 
    \item Specialized distillation that transfers knowledge at different layers (e.g., intermediate layer) and multiple stages is required to improve accuracy. 
    \item Small training data sizes make extreme compression difficult. 
\end{itemize}

\subsection{Is staged ternary-binary training necessary to mitigate the sharp performance drop?}

Previous works demonstrate that binary networks have more irregular loss surface than ternary models using curvature analysis. However, given that rough loss surface is a prevalent issue when training neural networks, we question whether all of them have a causal relationship with the difficulty in extreme compression.

We are first interested in the observation from previous work~\cite{bai-etal-2021-binarybert} that shows directly training a binarized network leads to a significant accuracy drop (e.g., up to 3.8\%) on GLUE in comparison to the accuracy loss from training a ternary network (e.g., up to 0.6\%). They suggest that this is because binary networks have a higher training loss and overall more complex loss surface. To overcome this sharp accuracy drop, they propose a staged ternary-binary training strategy in which one needs first to train a ternarized network and use a technique called weight splitting to use the trained ternary weights to initialize the binary model. 

It is not surprising that binarization leads to further accuracy drop as prior studies from computer vision also observe similar phenomenon~\cite{binarized-nn,how-to-trainbinary-nn}. However, prior works also identified that the difficulty of training binarized networks mostly resides in training with insufficient number of iterations or having too large learning rates, which can lead to frequent sign changes of the weights that make the learning of binarized networks unstable~\cite{how-to-trainbinary-nn}. Therefore, training a binarized network the same way as training a full precision network consistently achieves poor accuracy. Would increasing the training iterations and let the binarized network train longer under smaller learning rates help mitigate the performance drop from binarization?

To investigate this, we perform the following experiment: We remove the TernaryBERT training stage and weight splitting and directly train a binarized network. We use the same quantization schemes as in \cite{bai-etal-2021-binarybert}, e.g., a binary quantizer to quantize the weights: $w_i = \alpha \cdot sign(w_i), \alpha=\frac{1}{n}||w||_1$,
and a uniform symmetric INT8 quantizer to quantize the activations. We apply the One-Stage quantization-aware KD that will be explained in \sref{subsubsec:multi-stage-kd-analysis} to train the model. Nevertheless, without introducing the details of KD here, it does not affect our purpose of understanding the phenomena of a sharp performance drop since we only change the training iteration and learning rates while fixing other setups.
\begin{wraptable}{r}{9cm}
\centering
\caption{Different training budgets for the GLUE tasks.}
\label{table:budgets}
\begin{adjustbox}{width=0.99\linewidth}
\begin{tabular}{lcccc }
\toprule
\multirow{2}{*}{Dataset} & Data  &\multicolumn{3}{c}{Training epochs:} \\
                     & Aug.  &  Budget-A  &  Budget-B  &  Budget-C  \\\midrule
    QQP/MNLI & \xmark &  3  & 9  & 18 or 36 \\
    QNLI  &\cmark   &  1  & 3  &  6 or 9 \\
  SST-2/STS-B/RTE & \cmark &  1  & 3  &  12 \\
    CoLA/MRPC  & \cmark  &  1  & 3  &  12 or 18 \\
\bottomrule
\end{tabular}
\end{adjustbox}
\label{table:ta2}
\end{wraptable}

We consider three budgets listed in \tref{table:ta2}, which cover the practical scenarios of short, standard, and long training time, where Budget-A and Budget-B take into account the training details that appeared in BinaryBERT~\cite{bai-etal-2021-binarybert}, TernaryBERT~\cite{zhang-etal-2020-ternarybert}. Budget-C has a considerably larger training budget but smaller than TinyBert~\cite{bai-etal-2021-binarybert}. Meanwhile, we also perform a grid search of peak learning rates \{2e-5, 1e-4, 5e-4\}. For more training details on iterations and batch size per iteration, please see \tref{table:train-details}.

\begin{figure}[bt]
    \centering
    \includegraphics[width=1.\textwidth]{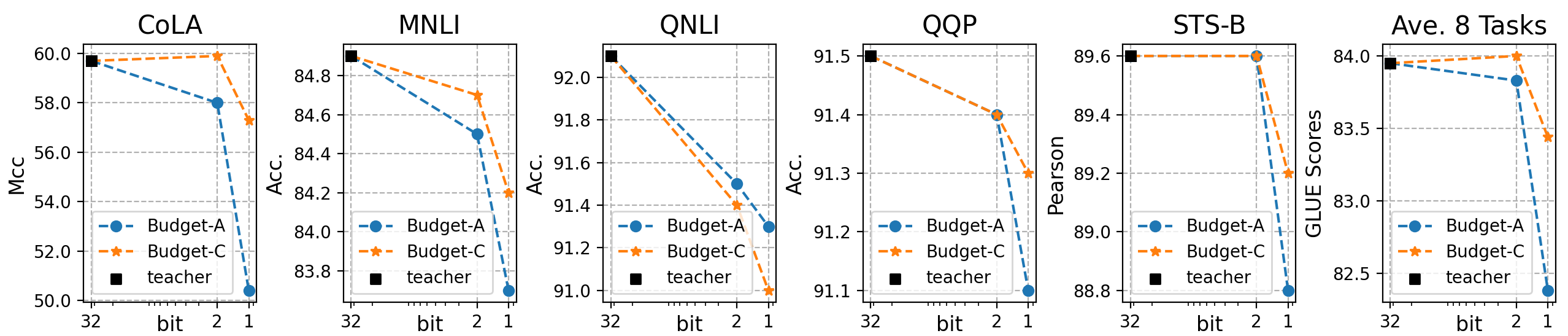}
    \caption{Performance of quantized \bertbase with different weight bits and 8-bit activation on the GLUE Benchmarks.
The results for orange and blue curves respectively represent the costs: (limited) Budget-A  and (sufficient) Budget-C. 
The fp32-teacher scores are shown by black square marker. }
    \label{fig:long-training}
    \vspace{-0.5cm}
\end{figure}

\begin{table}[ht]
\caption{\textbf{1-bit} quantization for \bertbase with various Budget-A, Budget-B and Budget-C.}
\centering
\label{table:1-bit-bert-longtraining}
\begin{adjustbox}{width=0.99\linewidth}
\centering
\begin{tabular}{llcccccccccccccc }
\toprule
\# & Cost      &  CoLA  & MNLI-m/-mm  &  MRPC      &  QNLI  &    QQP      & RTE    & SST-2 &   STS-B    & Avg. & Acc    \\
&        & Mcc     &   Acc/Acc   & F1/Acc     &  Acc   &  F1/Acc     & Acc    &  Acc  & Pear/Spea  &      &  Drop \\ 
\midrule
0 & Teacher         &  59.7  &  84.9/85.6   &  90.6/86.3  &  92.1  &  88.6/91.5   & 72.2   & 93.2  & 90.1/89.6 & 83.95 & -\\
\cdashline{1-12}
1 &Budget-A
              & 50.4  & 83.7/84.6  & 90.0/85.8  & 91.3  & 88.0/91.1  & 72.9  & 92.8  & 88.8/88.4  & 82.38 &  1.57 \\ 
2 &Budget-B
             & 55.6  & 84.1/84.4  & 90.4/86.0  & 90.8  & 88.3/91.3  & 72.6  & 93.1  & 88.9/88.5  & 82.98 &  0.97\\ 
3 &Budget-C
                & 57.3  & 84.2/84.4  & 90.7/86.5  & 91.0  & 88.3/91.3  & 74.0  & 93.1  & 89.2/88.8  & 83.44 & 0.51 \\ 
\bottomrule
\end{tabular}
\end{adjustbox}
\end{table}

We present our main results in \tref{table:1-bit-bert-longtraining} and \fref{fig:long-training} (see \tref{table:1-bit-bert-app} a full detailed results including three learning rates). We observe that although training binarized BERT with more iterations does not fully close the accuracy gap to the uncompressed full-precision model, the sharp accuracy drop from ternarization to binarization (the blue curve v.s. the orange curve in \fref{fig:long-training}) has largely been mitigated, leaving a much smaller accuracy gap to close. For example, when increasing the training time from Budget-A to Budget-C, the performance of CoLA boosts from 50.4 to 57.3, and the average score improves from 82.38 to 83.44 (\tref{table:1-bit-bert-longtraining}). These results indicate that previous studies~\cite{bai-etal-2021-binarybert} on binarized BERT were significantly under-trained. It also suggests that the observed sharp accuracy drop from binarization is primarily an optimization problem. We note that in the context of fine-tuning BERT models, Mosbach et al.~\cite{mosbach2021on} also observe that fine-tuning BERT can suffer from vanishing gradients, which causes the training converges to a "bad" valley with sub-optimal training loss. As a result, the authors also suggest increasing the number of iterations to train the BERT model to have stable and robust results. 

\begin{tcolorbox}
\textbf{Finding 1.} A longer training iterations with learning rate decay  is highly preferred for closing the accuracy gap of extreme quantization. 
\end{tcolorbox}

\textbf{Finding 1} seems natural. However, we argue that special considerations need to be taken for effectively doing extreme quantization before resorting to more complex solutions. 
\subsection{The role of multi-stage knowledge distillation}
\label{subsubsec:multi-stage-kd-analysis}
The above analysis shows that the previous binarized BERT models were severely undertrained. In this section, we further investigate the necessity of multi-stage knowledge distillation, which is quite complex because each stage has its own set of hyperparameters. Still, prior works claim to be crucial for improving the accuracy of extreme quantization. Investigating this problem is interesting because it can potentially admit a more straightforward solution with a cheaper cost. 

Prior work proposes an interesting knowledge distillation (KD), and here we call it \emph{Two-Stage} KD (2S-KD) \cite{jiao-etal-2020-tinybert}, which has been applied for extreme quantization~\cite{bai-etal-2021-binarybert}. The \twostage minimizes the losses of hidden states $\mathcal{L}_{\text{hidden}}$ and attention maps $\mathcal{L}_{\text{att}}$ and losses of prediction logits $\mathcal{L}_{\text{logit}}$ in two separate steps, where different learning rates are used in these two stages, e.g., a $\times 2.5$ larger learning rate is used for Stage-I than that for Stage-II. However, how to decide the learning rates and training epochs for these two stages is not well explained and will add additional hyperparameter tuning costs. We note that this strategy is very different from the deep knowledge distillation strategy used in~\cite{patient-kd,minilm-v2,sun2020mobilebert}, where knowledge distillation is performed with a single stage that minimizes the sum of the losses from prediction logits, hidden states, and attention maps. 
Despite the promising results in \cite{bai-etal-2021-binarybert}, the mechanism for why multi-stage KD improves accuracy is not well understood.

Generally, the knowledge distillation for transformer models can be formulated as minimizing the following objective:
\begin{align}
     \min_{\theta} \mathbb{E}_{x\sim D}[ \gamma \mathcal{L}_{\text{logit}}(x;\theta) + \beta(\mathcal{L}_{\text{att}}(x;\theta) + \mathcal{L}_{\text{hidden}}(x;\theta)] \label{eqn:objective} 
\end{align}
where $\mathcal{L}_{\text{logit}}$ denote the loss (e.g., KL divergence or mean square error) between student's and teacher's prediction logits, and $\mathcal{L}_{hidden}$ and $\mathcal{L}_{\text{att}}$) measures the loss of hidden states and attention maps. $\gamma\in\{0,1\}, \beta\in\{0,1\}$ are hyperparameters. See the detailed mathematical definition in \sref{sec:method-app}

To investigate the effectiveness of multi-stage, we compare three configurations (shown in \fref{fig:stage}):
\begin{itemize}
\item[(1)]\textbf{\onestage (One-stage KD):} $(\gamma, \beta) = (1,1)$ for $t\leq \mathcal{T}$;
\item[(2)]\textbf{\twostage (Two-stage KD):}  $(\gamma, \beta) = (0,1)$ if $t<\mathcal{T}/2$; $(\gamma, \beta) = (1,0)$ if $ \mathcal{T}/2 \leq t\leq \mathcal{T}$;
\item[(3)]\textbf{\threestage (Three-stage KD):} $(\gamma, \beta) = (0,1)$ if $t<\mathcal{T}/3$; $(\gamma, \beta) = (1,1)$ if $\mathcal{T}/3\leq t\leq 2\mathcal{T}/3$; $(\gamma, \beta) = (1,0)$ if $2\mathcal{T}/3\leq t\leq \mathcal{T}$.
\end{itemize} 
where $\mathcal{T}$ presents the total training budget and $t$ represents the training iterations. 

Notably, we created the \threestage, where we add a transition stage:  $\mathcal{L}_{\text{hidden}}+\mathcal{L}_{\text{att}}+\mathcal{L}_{\text{logit}}$ in between the first and second stage of the \twostage. The idea is to make the transition of the training objective smoother in comparison to the \twostage. In this case, the learning rate schedules will be correspondingly repeated in three times, and the peak learning rate of Stage-I is $\times 2.5$ higher than that of Stage-II and Stage-III. We apply all three KD methods independently  (under a fixed random seed) for binary quantization across different learning rates \{2e-5, 1e-4, 5e-4\} under the same training budget. We do not tune any other hyperparameters.

\tref{table:1-bit-bert-lr} shows the results under Budget-A (see \tref{table:1-bit-bert-app} for Budget-B/C). We note that when the learning rate is fixed to 2e-5 (i.e., Row 1, 5, and 9), the previously proposed \twostage~\cite{jiao-etal-2020-tinybert} (81.46) does show performance benefits over \onestage (80.33). Our newly created \threestage (81.57) obtains even better accuracy under the same learning rate. However, multi-stage makes the extreme compression procedure complex and inefficient because it needs to set the peak learning rates differently for different stages. Surprisingly, the simple \onestage easily outperforms both \twostage and \threestage  (Row 6 and 8) when we slightly increase the search space of the learning rates (e.g., 1e-4 and 5e-4). This means we can largely omit the burden of tuning learning rates and training epochs for each stage in multi-stage by using single-stage but with the same training iteration budget. 

\begin{tcolorbox}
\textbf{Finding 2.} Single-stage knowledge distillation with more training budgets and search of learning rates is sufficient to match or even exceed accuracy from multi-stage ones. 
\end{tcolorbox}

\begin{table}[ht]
\caption{\textbf{1-bit} quantization for BERT$_{\text{base}}$ with three different KD  under the training Budget-A.  }
\centering
\label{table:1-bit-bert-lr}
\begin{adjustbox}{width=0.99\linewidth}
\centering
\begin{tabular}{lcccccccccccccccc }
\toprule
\# & Stages    & learning    &  CoLA  & MNLI-m/-mm  &  MRPC      &  QNLI  &    QQP      & RTE    & SST-2 &   STS-B    & Avg.       \\
 &       &     rate &  Mcc     &   Acc/Acc   & F1/Acc     &  Acc   &  F1/Acc     & Acc    &  Acc & Pear/Spea  &  all      \\ 
\midrule
1  & \multirow{4}{*}{One-Stage}  &  2e-5  & 44.6  & 83.1/83.7  & 88.8/83.1  & 91.1  & 87.4/90.7  & 66.1  & 92.8  & 87.8/87.5  & 80.33 \\ 
2                       &                & 1e-4   & 50.4  & 83.7/84.6  & 90.1/85.3  & 91.3  & 88.0/91.1  & 71.5  & 92.7  & 88.8/88.4  & 82.16 \\ 
3                        &                & 5e-4   & 42.3  & 83.3/84.1  & 90.0/85.8  & 89.5  & 87.8/90.9  & 72.9  & 92.5  & 88.1/87.8  & 81.04 \\ \cdashline{3-14}
4                     &           &  Best (above)  & 50.4  & 83.7/84.6  & 90.0/85.8  & 91.3  & 88.0/91.1  & 72.9  & 92.8  & 88.8/88.4  & \textbf{82.38}  \\ \midrule
5          & \multirow{4}{*}{Two-Stage}  &  2e-5  & 48.2  & 83.3/83.8  & 89.3/84.6  & 90.7  & 87.7/90.9  & 70.4  & 92.5  & 88.7/88.4  & 81.46 \\ 
6                        &                & 1e-4   & 48.5  & 83.3/83.4  & 90.0/85.5  & 90.4  & 87.4/90.7  & 70.4  & 92.4  & 88.6/88.2  & 81.47  \\ 
7                         &                & 5e-4   & 16.2  & 74.9/76.4  & 89.7/84.8  & 87.8  & 85.0/89.0  & 68.2  & 91.6  & 86.2/86.5  & 75.01\\ \cdashline{3-14}
8                     &           &  Best (above)  & 48.5  & 83.3/83.8  & 90.0/85.5  & 90.7  & 87.7/90.9  & 70.4  & 92.5  & 88.7/88.4  & 81.59  \\\midrule
9          & \multirow{4}{*}{Three-Stage}  &  2e-5  & 49.3  & 83.1/83.6  & 89.5/84.3  & 91.0  & 87.6/90.9  & 70.8  & 92.4  & 88.7/88.3  & 81.57 \\ 
10                        &                & 1e-4   & 49.0  & 83.2/83.4  & 89.7/84.8  & 90.9  & 87.6/90.7  & 73.6  & 92.2  & 88.8/88.5  & 81.84 \\ 
11                        &                & 5e-4   & 29.5  & 81.3/81.9  & 89.1/83.6  & 88.4  & 85.0/89.3  & 66.1  & 91.9  & 84.1/83.9  & 77.34  \\ \cdashline{3-14}
12                     &           &  Best (above)  & 49.3  & 83.2/83.4  & 89.7/84.8  & 91.0  & 87.6/90.9  & 73.6  & 92.4  & 88.8/88.5  & 81.93 \\ 
\bottomrule
\end{tabular}
\end{adjustbox}
\vspace{-0.3cm}
\end{table}

\subsection{The importance of data augmentation}
Prior works augment the task-specific datasets using a text-editing technique by randomly replacing words in a sentence with their synonyms, based on their similarity measure on GloVe embeddings~\cite{jiao-etal-2020-tinybert}. They then use the augmented dataset for task-specific compression of BERT models, observing improved accuracy for extreme quantization~\cite{zhang-etal-2020-ternarybert,bai-etal-2021-binarybert}. 
They hypothesized that data augmentation (DA) is vital for compressing transformer models. 
However, as we have observed that the binarized networks were largely under-trained and had no clear advantage of multi-stage versus single-stage under a larger training budget,  it raises the question of the necessity of  DA.  

To better understand the importance of DA for extreme compression, we compare the end-task performance of both 1-bit quantized \bertbase models with and without data augmentation, based on previous findings in this paper that make extreme compression more robust. Table~\ref{table:da-1bit} shows results for the comparison of the 1-bit  \bertbase model under two different training budgets. Note that MNLI~/QQP does not use DA; we repeat the results. 
We find that regardless of whether shorter or longer iterations, removing DA leads to an average of 0.66 and 0.77 points drop in average GLUE score. Notably, the accuracy drops more than 1 point on smaller tasks such as CoLA, MPRC, RTE, STS-B. Furthermore,  
similar performance degradation is observed when removing DA from training an FP32 half-sized BERT model (more details will be introduced in  \sref{subsec:layer-reduction}), regardless of whether using \onestage or \twostage. These results indicate that DA is helpful, especially for extreme compressed models; DA significantly benefits from learning more diverse/small data to mitigate accuracy drops. 
\begin{table}[ht]
\caption{\textbf{The Comparison between the results with and without data augmentation (DA).} 
Row 1-4 is for \bertbase under 1-bit quantization using \onestage.  
Row 5-8 is for a layer-reduced \bertbase (six-layer)  under Budget-C without quantization (please see~\sref{subsec:layer-reduction} for more details).}
\label{table:da-1bit}
\begin{adjustbox}{width=0.99\linewidth}
\centering
\begin{tabular}{lccccccccccccccc }
\toprule
\# & Cost or & Data   &  CoLA  & MNLI-m/-mm  &  MRPC      &  QNLI  &    QQP      & RTE    & SST-2 &   STS-B    & Avg. & \multirow{2}{*}{DIFF}   \\
  &  Stages  & Aug. & Mcc     &   Acc/Acc   & F1/Acc     &  Acc   &  F1/Acc     & Acc    &  Acc  & Pear/Spea  &    all  &   \\   \midrule  

1 &  \multirow{2}{*}{Budget-A}  & \cmark   & 50.4  & 83.7/84.6  & 90.0/85.8  & 91.3  & 88.0/91.1  & 72.9  & 92.8  & 88.8/88.4  & 82.38 &\multirow{2}{*}{-0.77} \\ 
2&                               & \xmark   & 48.9  & 83.7/84.6  & 89.0/83.6  & 91.0  & 88.0/91.1  & 71.5  & 92.5  & 87.6/87.5  & 81.61 & \\ \cdashline{1-13} 
3 &  \multirow{2}{*}{Budget-C}  & \cmark  & 57.3  & 84.2/84.4  & 90.7/86.5  & 91.0  & 88.3/91.3  & 74.0  & 93.1  & 89.2/88.8  & 83.44 & \multirow{2}{*}{-0.66} \\  
4&                               & \xmark  & 55.0  & 84.2/84.4  & 90.0/85.0  & 90.7  & 88.3/91.3  & 73.3  & 92.9  & 88.2/87.9  & 82.78 &  \\ \midrule  




5 &  \multirow{2}{*}{One-stage }  & \cmark    & 56.9  & 84.5/84.9  & 89.7/84.8  & 91.7  & 88.5/91.5  & 71.8  & 93.5  & 89.8/89.4  & 83.27 & \multirow{2}{*}{-2.18} \\
6&                               & \xmark      & 52.2  & 84.5/84.9  & 88.4/82.4  & 91.4  & 88.5/91.5  & 63.9  & 92.9  & 86.5/86.3  & 81.09 &  \\ \cdashline{1-13}
7 &  \multirow{2}{*}{Two-stage}  & \cmark     & 56.0  & 84.1/84.2  & 89.7/84.8  & 91.5  & 88.2/91.3  & 71.8  & 93.3  & 89.4/89.1  & 82.93 & \multirow{2}{*}{-2.03}  \\  
8&                               & \xmark     & 50.2  & 84.1/84.2  & 88.7/83.1  & 90.8  & 88.2/91.3  & 64.6  & 92.9  & 86.9/86.7  & 80.90 &  \\  
 \bottomrule
\end{tabular}
\end{adjustbox}
\vspace{-0.3cm}
\end{table}
\begin{tcolorbox}
\textbf{Finding 3.} Training without DA hurts performance on downstream tasks for various compression tasks, especially on smaller tasks. 
\end{tcolorbox}

We remark that although our conclusion for DA is consistent with the finding in \cite{bai-etal-2021-binarybert}, we have performed a far more well-through experiment than \cite{bai-etal-2021-binarybert} where they consider a single budget and \twostage only.

\section{Interplay of KD, Long Training, Data Augmentation and Layer Reduction}\label{subsec:layer-reduction}

In the previous section, we found strong evidence for statistically significant benefits from \onestage compared to \twostage and \threestage. Moreover, a considerable long training can greatly lift the score to the level same as the full precision teacher model. Although 1-bit \bertbase already achieves $\times32$ smaller size than its full-precision counterpart, a separate line of research focuses on changing the model architecture to reduce the model size. Notably, one of the promising methods -- reducing model sizes via knowledge distillation -- has long been studied and shown good performances  \cite{Fan2020Reducing,zhang2020accelerating, distill-bert,minilm-v2,sun2020mobilebert}. 

While many works focus on innovating better knowledge distillation methods for higher compression ratio and better accuracy, given our observations in Section~\ref{subsec:hypothesis-analysis} that extreme quantization can effectively reduce the model size,  we are interested in \emph{to what extent can extreme quantization benefit state-of-the-art distilled models?}

To investigate this issue, we perform the following experiments: We prepare five student models, all having 6 layers with the same hidden dimension 768: (1) a pretrained TinyBERT$_6$ \cite{jiao-etal-2020-tinybert}\footnote{The checkpoint of TinyBert is downloaded from  \href{https://huggingface.co/huawei-noah/TinyBERT_General_6L_768D/tree/main}{their uploaded huggingface.co}.}; (2) MiniLMv2$_6$ \cite{minilm-v2}\footnote{The checkpoint of miniLMv2 is from \href{https://1drv.ms/u/s!AjHn0yEmKG8qiyLmvLxXOSgpTxxm}{their github}.}; (3) Top-BERT$_6$: using the top 6 layers of the fine-tuned BERT-base model to initialize the student model; (4) 
Bottom-BERT$_6$: using the bottom six layers of the fine-tuned BERT-base model to initialize the student; (5) Skip-BERT$_6$: using every other layer of the fine-tuned BERT-based model to initialize the student. In all cases, we fine-tune \bertbase for each task as the teacher model (see the teachers' performance in \tref{table:pretrain}, Row 0). We choose TinyBERT and MiniLM because they are the state-of-the-art for knowledge distillation of BERT models. We choose the other three configurations because prior work~\cite{patient-kd,greedy-layer-reduction} also suggested that layer pruning is also a practical approach for task-specific compression. 
For the initialization of the student model, both TinyBERT$_6$ and MiniLM$_6$ are initialized with weights distilled from \bertbase through pretraining distillation, and the other three students (e.g., top, bottom, skip) are obtained from the fine-tuned teacher model \emph{without incurring any pretraining training cost}. 
\begin{table}[ht]
\caption{Pre-training does not show benefits for layer reduction. 
Row 3 (rep.*) is a reproduced result by following the training recipe in \cite{jiao-etal-2020-tinybert}. }\label{table:pretrain}
\begin{adjustbox}{width=0.99\linewidth}
\centering
\begin{tabular}{llccccccccccccccc}
\toprule
\# & Model  &  size    &  CoLA  & MNLI-m/-mm  &  MRPC      &  QNLI  &    QQP      & RTE    & SST-2 &   STS-B    & Avg. & Acc.    \\
 &   &      &  Mcc     &   Acc/Acc   & F1/Acc     &  Acc   &  F1/Acc     & Acc    &  Acc  & Pear/Spea  &   all   & Drop   \\   \midrule    
1 & BERT-{base} fp32             & 417.2   &  59.7  &  84.9/85.6   &  90.6/86.3  &  92.1  &  88.6/91.5   & 72.2   & 93.2  & 90.1/89.6 & 83.95 & -\\\midrule
\multicolumn{13}{l}{\textbf{Training cost: greater than Budget-C (see \cite{jiao-etal-2020-tinybert} or \sref{sec:method-app})}}\\
2 & Pretrained TinyBERT$_{6}$ (\cite{jiao-etal-2020-tinybert}) & 255.2 ($\times 1.6$)   &  54.0  &  84.5/84.5   &  90.6/86.3  &  91.1  &  88.0/91.1   & 73.4   & 93.0  & 90.1/89.6 & 83.11 & -0.84\\
3 & Pretrained TinyBERT$_{6}$ (rep.*)  & 255.2 ($\times 1.6$)   &  56.9  &  84.4/84.8   &  90.1/85.5  &  91.3  &  88.4/91.4   & 72.2   & 93.2  &  90.3/90.0   & 83.33 & -0.62 \\\midrule
\multicolumn{13}{l}{\textbf{Training cost: Budget-C}}\\
4 &Pretrained  TinyBERT$_{6}$    & 255.2 ($\times 1.6$) & 54.4 & 84.6/84.3 & 90.4/86.3 & 91.5 & 88.5/91.5 & 69.7 & 93.3 & 89.2/89.0   & 82.76 & -1.19\\
5 & Pretrained MiniLM$_6$-v2        & 255.2 ($\times 1.6$) & 55.4 & 84.4/84.5 & 90.7/86.5 & 91.4 & 88.5/91.5 & 71.8 & 93.3 & 89.4/89.0   & 83.13 & -0.82 \\
6 &Skip-BERT$_6$   (ours)    & 255.2 ($\times 1.6$) & 56.9 & 84.6/84.9 & 90.4/85.8 & 91.8 & 88.6/91.6 & 72.6 & 93.5 & 89.8/89.4 & \textbf{83.50} & -0.45 \\   \cdashline{1-13}
7 &Skip-BERT$_5$   (ours)    & 228.2 ($\times 1.8$) & 57.9 & 84.3/85.1 & 90.1/85.5 & 91.4 & 88.5/91.5 & 72.2 & 93.3 & 89.2/88.9 & \textbf{83.38} &-0.57 \\
8 &Skip-BERT$_4$   (ours)   & 201.2 ($\times 2.1$)  & 53.3 & 83.2/83.4 & 90.0/85.3   & 90.8 & 88.2/91.3 & 70.0 & 93.5 & 88.8/88.4 & 82.18 & -1.77 \\
 \bottomrule
\end{tabular}
\end{adjustbox}
\end{table}
We apply \onestage as we also verify that \twostage and \threestage are under-performed (See \tref{table:stages-fp32} in the appendix). We set Budget-C as our training cost because the training budget in reproducing the results for TinyBert is greatly larger than Budget-C illustrated above. We report their best validation performances in \tref{table:pretrain} across the three learning rate \{5e-5, 1e-4, 5e-4\}.  
We apply \onestage and budget-C for this experiment. We report their best validation performances in \tref{table:pretrain} across three learning rates \{5e-5, 1e-4, 5e-4\}. For complete statistics with these learning rates, please refer to  \tref{table:pretrain-appendix}. We make a few observations:

First, perhaps a bit surprisingly, the results in \tref{table:pretrain} show that there are no significant improvements from using the more computationally expensive pre-pretraining distillation in comparison to lightweight layer reduction: Skip-BERT$_6$ (Row 7) achieves the highest average score 83.50. This scheme achieves the highest score on larger and more robust tasks such as MNLI/QQP/QNLI/SST-2. 

One noticeable fact in \tref{table:pretrain} is that under the same budget (Budget-C), the accuracy drop of Skip-BERT$_6$ is about 0.74 (0.37) higher than TinyBERT$_6$ in Row 5 (MINILM$_6$-v2 in Row 6). Note that layer reduction without pretraining has also been addressed in \cite{poorstable-fine-tune-bert-man-bert}. However, their KD is limited to logits without DA, and the performance is not better a trained DistillBERT.

Second, with the above encouraging results from a half-size model (Skip-BERT$_6$), we squeeze the depths into five and four layers. The five-/four-layer student is initialized from  $\ell$-layer of teacher with $\ell\in \{3,5,7,9,11\}$ or $\ell\in \{3,6,9,12\}$. We apply the same training recipe as Skip-BERT$_6$ and report the results in Row 8 and 9 in \tref{table:pretrain}. There is little performance degradation in our five-layer model (83.28) compared to its six-layer counterpart (83.50); Interestingly, CoLA and MNLI-mm even achieve higher accuracy with smaller model sizes. Meanwhile, when we perform even more aggressive compression by reducing the depth to 4, the result is less positive as the average accuracy drop is around 1.3 points. 

Third, among three lightweight layer reduction methods, we confirm that a Skip-\# student performs better than those using Top-\# and Bottom-\#. We report the full results of this comparison in the Appendix \tref{table:pretrain-student-slection}. This observation is consistent with \cite{jiao-etal-2020-tinybert}. However, their layerwise distillation method is not the same as ours, e.g, they use the non-adapted pretrained BERT model to initialize the student, whereas we use the fine-tuned BERT weights. We remark that our finding is in contrast to \cite{poorstable-fine-tune-bert-man-bert} which is perhaps because the KD in \cite{poorstable-fine-tune-bert-man-bert} only uses logits distillation without layerwise distillation.
\begin{tcolorbox}
\textbf{Finding 4.} Lightweight layer reduction matches or even exceeds expensive pre-training distillation for task-specific compression.
\end{tcolorbox}
\section{Proposed Method for Further Pushing the Limit of Extreme Compression}
\label{sec:design}

Based on our studies, we propose a simple yet effective method tailored for extreme lightweight compression. \fref{fig:propose-method} (right) illustrates our proposed method: \OURS, which consists of only 2 steps:

\textbf{Step I: Lightweight layer reduction.} Unlike the common layer reduction method where the layer-reduced model is obtained through computationally expensive pre-training distillation, we select a subset of the fine-tuned teacher weights as a lightweight layer reduction method (e.g., either through simple heuristics as described in Section~\ref{subsec:layer-reduction} or search-based algorithm as described in \cite{greedy-layer-reduction}) to initialize the layer-reduced model. When together with the other training strategies identified in this paper, we find that such a lightweight scheme allows for achieving a much larger compression ratio while setting a new state-of-the-art result compared to other existing methods. 

\textbf{Step II: 1-bit quantization by applying \onestage with DA and long training.} 
Once we obtain the layer-reduced model, we apply the quantize-aware \onestage, proven to be the most effective in \sref{subsubsec:multi-stage-kd-analysis}. To be concrete, we use an ultra-low bit (1-bit/2-bit) quantizer to compress the layer-reduced model weights for a forward pass and then use STE during the backward pass for passing gradients. Meanwhile, we minimize the single-stage deep knowledge distillation objective with data augmentation enabled and longer training Budget-C (such that the training loss is close to zero).  

\begin{table}[ht]
\caption{1-/2-bit quantization of the layer-reduced model. 
The last column (Acc. drop) is the accuracy drop from their own fp-32 models. 
See full details in~\tref{table:quantization-smaller-bert-app}.}
\label{table:quantization-smaller-bert}
\begin{adjustbox}{width=0.99\linewidth}
\centering
\begin{tabular}{lclccccccccccccc }
\toprule
 & &    bit & size    &  CoLA  & MNLI-m/-mm  &  MRPC      &  QNLI  &    QQP      & RTE    & SST-2 &   STS-B    & Avg. & Acc.   \\
\#& Method &(\#-layer)   &   (MB) & Mcc     &   Acc/Acc   & F1/Acc     &  Acc   &  F1/Acc     & Acc    &  Acc  & Pear/Spea  &   all   & drop \\   \midrule    
1 &        \cite{zhang-etal-2020-ternarybert}                                                           &2 (6L)  & 16.0 ($\times 26.2$)  & 53.0  & 83.4/83.8  & 91.5/88.0  & 89.9  & 87.2/90.5  & 71.8  & 93.0  & 86.9/86.5  & 82.26 & -0.76 \\    \midrule

2&\multirow{5}{*}{Ours}  &  {2 (6L) }    & 16.0 ($\times 26.2$)  & 53.8  & 83.6/84.2  & 90.5/86.3  & 90.6  & 88.2/91.3  & 73.6  & 93.6  & 89.0/88.7 & \textbf{82.89} & -0.44 \\  
3&                      &  {2 (5L) }  & 14.2 ($\times 29.3$) & 53.9  & 83.3/84.1  & 90.4/86.0  & 90.4  & 88.2/91.2  & 71.8  & 93.0  & 88.4/88.0  & \textbf{82.46} & -0.72 \\
4&                      &  {2 (4L)}  & 12.6 ($\times 33.2$)  & 50.3  & 82.5/83.0  & 90.0/85.3  & 89.2  & 87.8/91.0  & 69.0  & 92.8  & 87.9/87.4  & 81.22 & -0.90 \\ \cdashline{3-14}

5&                       & {1 (6L) }    & 8.0 ($\times 52.3$)    & 52.3  & 83.4/83.8  & 90.0/85.3  & 89.4  & 87.9/91.1  & 68.6  & 93.1  & 88.4/88.0  & 81.71 & -1.51 \\  
6&                      &{1 (5L) }  &7.1 ($\times 58.5$) &  52.2  & 82.9/83.2  & 89.9/85.0  & 88.5  & 87.6/90.8  & 69.3  & 92.9  & 87.3/87.0  & 81.34 & -1.84 \\  

7&                    &  {1 (4L)}  & 6.3 ($\times 66.4$)  & 48.3  & 82.0/82.3  & 89.9/85.5  & 87.7  & 86.9/90.4  & 63.9  & 92.4  & 87.1/86.7  & 79.96 & -2.01 \\ 
 \bottomrule
\end{tabular}
\end{adjustbox}
\vspace{-0.3cm}
\end{table}
\paragraph{Evaluation Results.} The results of our approach are presented in \tref{table:quantization-smaller-bert}, which we include multiple layer-reduced models with both 1-bit and 2-bit quantization. We make the following two observations: \\
\textbf{(1)} For the 2-bit + 6L model (Row 1 and Row 2), 
\OURS achieves  0.63 points higher accuracy than 2-bit quantized model TernaryBERT \cite{zhang-etal-2020-ternarybert}. This remarkable improvement consists of two factors: (a) our fp-32 layer-reduced model is better (b) when binarization, we train longer under three learning rate searches. To understand how much the factor-(b) benefits, we may refer to the accuracy drop from their fp32 counterparts (last col.) where ours only drops 0.44 points and TinyBERT$_6$ drops 0.76.\\
\textbf{(2)}   When checking between our 2-bit  five-layer (Row 3, 82.46) in \tref{table:quantization-smaller-bert}  and 2-bit TinyBERT$_6$ (Row 1, 82.26), our quantization is much better, while our model size is 12.5\% smaller, setting a new state-of-the-art result for 2-bit quantization with this size.\\ \textbf{(3)}  Let us take a further look at accuracy drops (last column) for six/five/four layers, which are 0.44 (1.51), 0.72 (1.84), and 0.9 (2.01) respectively. It is clear that smaller models become much more brittle for extreme quantization than \bertbase, especially the 1-bit compression as the degradation is $\times$3  or $\times$4 higher than the 2-bit quantization.   

 Besides the reported results above,  we have also done a well-thorough investigation on the effectiveness of adding a low-rank (LoRa) full-precision weight matrices to see if it will further push the accuracy for the 1-bit layer-reduced models. However, we found a negative conclusion on using LoRa. Interestingly, if we repeat another training with the 1-bit quantization after the single-stage long training, there is another ~0.3 increase in the average GLUE score (see \tref{table:continue-1bit}). Finally, we also verify that prominent teachers can indeed help to improve the results (see appendix). 
\vspace{-0.3cm}
\section{Conclusions}
\label{sec:conclusions}
We carefully design and perform extensive experiments to investigate the contemporary existing extreme quantization methods \cite{jiao-etal-2020-tinybert, bai-etal-2021-binarybert} by fine-tuning pre-trained \bertbase models with various training budgets and learning rate search. Unlike \cite{bai-etal-2021-binarybert}, we find that there is no sharp accuracy drop if long training with data augmentation is used and that multi-stage KD and pretraining introduced in \cite{jiao-etal-2020-tinybert} is not a must in our setup. Based on the finding, we derive a user-friendly celebrating recipe for extreme quantization (see \fref{fig:propose-method}, right), which allows us to achieve a larger compression ratio and higher accuracy. See our summarized results in \tref{table:ours-summary}.\footnote{We decide not to include many other great works \cite{lan2019albert,zadeh2020gobo,distill-bert, qin2022bibert} since they do not use data augmentation or their setups are not close. }

\begin{wraptable}{r}{8cm}
\centering
\caption{Summary of task-specific performance of MNLI and GLUE scores. 
Also see~\fref{fig:sota}. }
\centering
\label{table:ours-summary}
\begin{adjustbox}{width=0.99\linewidth}
\centering
\begin{tabular}{lllccc }
\toprule
\# & Model (8-bit activation)   &  size (MB)     & MNLI-m/-mm  & GLUE Score       \\
\midrule
1 & \bertbase (Teacher)	       & 417.2 ($\times 1.0$) & 84.9/85.6  & 83.95  \\\midrule
2 &  TinyBERT$_{6}$ (rep*) & 255.2 ($\times 1.6$) & 84.4/84.8  & 83.33  \\
3 &  \OURS-BERT$_6$ (ours) & 255.2 ($\times 1.6$) & \textbf{84.6}/84.9  &  \textbf{83.50}  \\
4 & \OURS-BERT$_5$ (ours) & 228.2 ($\times 1.8$) & 84.3/\textbf{85.1}  & \textbf{83.38}  \\\midrule
5 & 2-bit BERT~\cite{zhang-etal-2020-ternarybert} & 26.8 ($\times 16.0$)   & 83.3/83.3  &   82.73         \\
6 & 2-bit \OURS-BERT (ours) & 26.8 ($\times 16.0$)   & \textbf{84.6/84.7} &  \textbf{84.10}  \\\cdashline{1-5}
7 &1-bit BERT (TWN)~\cite{bai-etal-2021-binarybert} & 16.5 ($\times 25.3$)   & 84.2/ \textbf{84.7}  &   82.49         \\
8 &1-bit BERT (BWN)~\cite{bai-etal-2021-binarybert} & 13.4 ($\times 32.0$)   & 84.2/84.0  &   82.29      \\
9 &1-bit \OURS-BERT (ours) & 13.4 ($\times 32.0$)  &  84.2/84.4    &  \textbf{83.44}   \\ \midrule
10 &2-bit TinyBERT$_{6}$~\cite{zhang-etal-2020-ternarybert} & 16.0 ($\times 26.2$)  & 83.4/83.8  &  82.26   \\ 
11 &2-bit \OURS-BERT$_{6}$ (ours) & 16.0 ($\times 26.2$)  &  \textbf{83.6/84.2}  &  \textbf{82.89}   \\ 
12 &2-bit \OURS-BERT$_{5}$ (ours) & 14.2 ($\times 29.3$)  & 83.3/84.1  &  \textbf{82.46}  \\ \cdashline{1-5}
13 &1-bit \OURS-BERT$_{6}$ (ours) & 8.0 ($\times 52.3$)  & 83.4/83.8  &  81.71  \\ 
14 &1-bit \OURS-BERT$_{5}$ (ours) & 7.1 ($\times 58.5$)  & 82.9/83.2  &  81.34  \\ 
\bottomrule
\end{tabular}
\end{adjustbox}

\end{wraptable}

\textbf{Discussion and future work.}
Despite our encouraging results, we note that there is a caveat in our \textbf{Step I} when naively applying our method (i.e., initializing the student model as a subset of the teacher) to reduce the width of the model as the weight matrices dimension of teacher do not fit into that of students. We found that the accuracy drop is considerably sharp. Thus, in this domain, perhaps pertaining distillation~\cite{jiao-etal-2020-tinybert} could be indeed useful, or Neural Architecture Search (AutoTinyBERT~\cite{yin2021autotinybert}) could be another promising direction. We acknowledge that, in order to make a  fair comparison with \cite{zhang-etal-2020-ternarybert,bai-etal-2021-binarybert},  our investigations are based on the classical 1-bit~\cite{rastegari2016xnor} and 2-bit~\cite{li2016ternary} algorithms and without quantization on layernorm~\cite{ba2016layer}.

Exploring the benefits of long training on an augmented dataset with different 1-/2-bit algorithms  (with or without quantization on layer norm) would be of potential interests~\cite{kim2021bert, qin2022bibert}.
Finally, as our experiments focus on \bertbase model, future work can be understanding how our conclusion transfers to decoder models such as GPT-2/3~\cite{gpt-2,gpt-3}.

\section*{Acknowledgments}
This work is done within the DeepSpeed team in Microsoft. We appreciate the help from the DeepSpeed team members. Particularly, we thank Jeff Rasley for coming up with the name of our method and Elton Zheng for solving the engineering issues. We thank the engineering supports from the Turing team in Microsoft.
\clearpage

{
\bibliographystyle{plain}
\bibliography{ref.bib}

\begin{thebibliography}{10}

\bibitem{kd-internal-amazon}
Gustavo Aguilar, Yuan Ling, Yu~Zhang, Benjamin Yao, Xing Fan, and Chenlei Guo.
\newblock Knowledge distillation from internal representations.
\newblock In {\em The Thirty-Fourth {AAAI} Conference on Artificial
  Intelligence, {AAAI} 2020}, pages 7350--7357. {AAAI} Press, 2020.

\bibitem{ba2016layer}
Jimmy~Lei Ba, Jamie~Ryan Kiros, and Geoffrey~E Hinton.
\newblock Layer normalization.
\newblock {\em arXiv preprint arXiv:1607.06450}, 2016.

\bibitem{bai-etal-2021-binarybert}
Haoli Bai, Wei Zhang, Lu~Hou, Lifeng Shang, Jin Jin, Xin Jiang, Qun Liu,
  Michael Lyu, and Irwin King.
\newblock {B}inary{BERT}: Pushing the limit of {BERT} quantization.
\newblock In {\em Proceedings of the 59th Annual Meeting of the Association for
  Computational Linguistics and the 11th International Joint Conference on
  Natural Language Processing (Volume 1: Long Papers)}, pages 4334--4348,
  Online, August 2021. Association for Computational Linguistics.

\bibitem{banner2018scalable}
Ron Banner, Itay Hubara, Elad Hoffer, and Daniel Soudry.
\newblock Scalable methods for 8-bit training of neural networks.
\newblock {\em Advances in neural information processing systems}, 31, 2018.

\bibitem{bengio2013estimating}
Yoshua Bengio, Nicholas L{\'e}onard, and Aaron Courville.
\newblock Estimating or propagating gradients through stochastic neurons for
  conditional computation.
\newblock {\em arXiv preprint arXiv:1308.3432}, 2013.

\bibitem{gpt-3}
Tom~B. Brown, Benjamin Mann, Nick Ryder, Melanie Subbiah, Jared Kaplan,
  Prafulla Dhariwal, Arvind Neelakantan, Pranav Shyam, Girish Sastry, Amanda
  Askell, Sandhini Agarwal, Ariel Herbert{-}Voss, Gretchen Krueger, Tom
  Henighan, Rewon Child, Aditya Ramesh, Daniel~M. Ziegler, Jeffrey Wu, Clemens
  Winter, Christopher Hesse, Mark Chen, Eric Sigler, Mateusz Litwin, Scott
  Gray, Benjamin Chess, Jack Clark, Christopher Berner, Sam McCandlish, Alec
  Radford, Ilya Sutskever, and Dario Amodei.
\newblock Language models are few-shot learners.
\newblock {\em CoRR}, abs/2005.14165, 2020.

\bibitem{cer2017semeval}
Daniel Cer, Mona Diab, Eneko Agirre, Inigo Lopez-Gazpio, and Lucia Specia.
\newblock Semeval-2017 task 1: Semantic textual similarity-multilingual and
  cross-lingual focused evaluation.
\newblock {\em arXiv preprint arXiv:1708.00055}, 2017.

\bibitem{bert-lottery}
Tianlong Chen, Jonathan Frankle, Shiyu Chang, Sijia Liu, Yang Zhang, Zhangyang
  Wang, and Michael Carbin.
\newblock The lottery ticket hypothesis for pre-trained bert networks.
\newblock {\em Advances in neural information processing systems},
  33:15834--15846, 2020.

\bibitem{dagan2013recognizing}
Ido Dagan, Dan Roth, Mark Sammons, and Fabio~Massimo Zanzotto.
\newblock Recognizing textual entailment: Models and applications.
\newblock {\em Synthesis Lectures on Human Language Technologies}, 6(4):1--220,
  2013.

\bibitem{bert}
Jacob Devlin, Ming{-}Wei Chang, Kenton Lee, and Kristina Toutanova.
\newblock {BERT:} pre-training of deep bidirectional transformers for language
  understanding.
\newblock In {\em Proceedings of the 2019 Conference of the North American
  Chapter of the Association for Computational Linguistics: Human Language
  Technologies, {NAACL-HLT} 2019)}, pages 4171--4186, 2019.

\bibitem{dolan2005automatically}
William~B Dolan and Chris Brockett.
\newblock Automatically constructing a corpus of sentential paraphrases.
\newblock In {\em Proceedings of the Third International Workshop on
  Paraphrasing (IWP2005)}, 2005.

\bibitem{dong2019hawq}
Zhen Dong, Zhewei Yao, Amir Gholami, Michael~W Mahoney, and Kurt Keutzer.
\newblock Hawq: Hessian aware quantization of neural networks with
  mixed-precision.
\newblock In {\em Proceedings of the IEEE/CVF International Conference on
  Computer Vision}, pages 293--302, 2019.

\bibitem{Fan2020Reducing}
Angela Fan, Edouard Grave, and Armand Joulin.
\newblock Reducing transformer depth on demand with structured dropout.
\newblock In {\em International Conference on Learning Representations}, 2020.

\bibitem{gou2021knowledge}
Jianping Gou, Baosheng Yu, Stephen~J Maybank, and Dacheng Tao.
\newblock Knowledge distillation: A survey.
\newblock {\em International Journal of Computer Vision}, 129(6):1789--1819,
  2021.

\bibitem{deberta}
Pengcheng He, Xiaodong Liu, Jianfeng Gao, and Weizhu Chen.
\newblock Deberta: Decoding-enhanced bert with disentangled attention.
\newblock In {\em International Conference on Learning Representations}, 2020.

\bibitem{kd-hinton}
Geoffrey~E. Hinton, Oriol Vinyals, and Jeffrey Dean.
\newblock Distilling the knowledge in a neural network.
\newblock {\em CoRR}, abs/1503.02531, 2015.

\bibitem{hou2020dynabert}
Lu~Hou, Zhiqi Huang, Lifeng Shang, Xin Jiang, Xiao Chen, and Qun Liu.
\newblock Dynabert: Dynamic bert with adaptive width and depth.
\newblock {\em Advances in Neural Information Processing Systems},
  33:9782--9793, 2020.

\bibitem{binarized-nn}
Itay Hubara, Matthieu Courbariaux, Daniel Soudry, Ran El{-}Yaniv, and Yoshua
  Bengio.
\newblock Binarized neural networks.
\newblock In Daniel~D. Lee, Masashi Sugiyama, Ulrike von Luxburg, Isabelle
  Guyon, and Roman Garnett, editors, {\em Advances in Neural Information
  Processing Systems 29: Annual Conference on Neural Information Processing
  Systems 2016, December 5-10, 2016, Barcelona, Spain}, pages 4107--4115, 2016.

\bibitem{iyer2017first}
Shankar Iyer, Nikhil Dandekar, and Kornl Csernai.
\newblock First quora dataset release: Question pairs, 2017.
\newblock {\em URL https://data. quora.
  com/First-Quora-Dataset-Release-Question-Pairs}, 2017.

\bibitem{jiao-etal-2020-tinybert}
Xiaoqi Jiao, Yichun Yin, Lifeng Shang, Xiao Chen, Linlin Li, Fang Wang, and Qun
  Liu.
\newblock {T}iny{BERT}: Distilling {BERT} for natural language understanding.
\newblock In {\em Findings of the Association for Computational Linguistics:
  EMNLP 2020}, pages 4163--4174, Online, November 2020. Association for
  Computational Linguistics.

\bibitem{is-bert-robust}
Di~Jin, Zhijing Jin, Joey~Tianyi Zhou, and Peter Szolovits.
\newblock Is {BERT} really robust? {A} strong baseline for natural language
  attack on text classification and entailment.
\newblock In {\em The Thirty-Fourth {AAAI} Conference on Artificial
  Intelligence}, pages 8018--8025. {AAAI} Press, 2020.

\bibitem{juefei2017local}
Felix Juefei-Xu, Vishnu Naresh~Boddeti, and Marios Savvides.
\newblock Local binary convolutional neural networks.
\newblock In {\em Proceedings of the IEEE conference on computer vision and
  pattern recognition}, pages 19--28, 2017.

\bibitem{kim2016bitwise}
Minje Kim and Paris Smaragdis.
\newblock Bitwise neural networks.
\newblock {\em arXiv preprint arXiv:1601.06071}, 2016.

\bibitem{kim2021bert}
Sehoon Kim, Amir Gholami, Zhewei Yao, Michael~W Mahoney, and Kurt Keutzer.
\newblock I-bert: Integer-only bert quantization.
\newblock In {\em International conference on machine learning}, pages
  5506--5518. PMLR, 2021.

\bibitem{lagunas2021block}
Fran{\c{c}}ois Lagunas, Ella Charlaix, Victor Sanh, and Alexander~M Rush.
\newblock Block pruning for faster transformers.
\newblock In {\em Proceedings of the 2021 Conference on Empirical Methods in
  Natural Language Processing}, pages 10619--10629, 2021.

\bibitem{lan2019albert}
Zhenzhong Lan, Mingda Chen, Sebastian Goodman, Kevin Gimpel, Piyush Sharma, and
  Radu Soricut.
\newblock Albert: A lite bert for self-supervised learning of language
  representations.
\newblock {\em arXiv preprint arXiv:1909.11942}, 2019.

\bibitem{li2016ternary}
Fengfu Li, Bo~Zhang, and Bin Liu.
\newblock Ternary weight networks.
\newblock {\em arXiv preprint arXiv:1605.04711}, 2016.

\bibitem{roberta}
Yinhan Liu, Myle Ott, Naman Goyal, Jingfei Du, Mandar Joshi, Danqi Chen, Omer
  Levy, Mike Lewis, Luke Zettlemoyer, and Veselin Stoyanov.
\newblock Roberta: {A} robustly optimized {BERT} pretraining approach.
\newblock {\em CoRR}, abs/1907.11692, 2019.

\bibitem{liu2021post}
Zhenhua Liu, Yunhe Wang, Kai Han, Wei Zhang, Siwei Ma, and Wen Gao.
\newblock Post-training quantization for vision transformer.
\newblock {\em Advances in Neural Information Processing Systems}, 34, 2021.

\bibitem{ladabert}
Yihuan Mao, Yujing Wang, Chufan Wu, Chen Zhang, Yang Wang, Quanlu Zhang, Yaming
  Yang, Yunhai Tong, and Jing Bai.
\newblock Ladabert: Lightweight adaptation of {BERT} through hybrid model
  compression.
\newblock In Donia Scott, N{\'{u}}ria Bel, and Chengqing Zong, editors, {\em
  Proceedings of the 28th International Conference on Computational
  Linguistics, {COLING} 2020, Barcelona, Spain (Online), December 8-13, 2020},
  pages 3225--3234. International Committee on Computational Linguistics, 2020.

\bibitem{mellempudi2017ternary}
Naveen Mellempudi, Abhisek Kundu, Dheevatsa Mudigere, Dipankar Das, Bharat
  Kaul, and Pradeep Dubey.
\newblock Ternary neural networks with fine-grained quantization.
\newblock {\em arXiv preprint arXiv:1705.01462}, 2017.

\bibitem{mosbach2021on}
Marius Mosbach, Maksym Andriushchenko, and Dietrich Klakow.
\newblock On the stability of fine-tuning {\{}bert{\}}: Misconceptions,
  explanations, and strong baselines.
\newblock In {\em International Conference on Learning Representations}, 2021.

\bibitem{nagel2020up}
Markus Nagel, Rana~Ali Amjad, Mart Van~Baalen, Christos Louizos, and Tijmen
  Blankevoort.
\newblock Up or down? adaptive rounding for post-training quantization.
\newblock In {\em International Conference on Machine Learning}, pages
  7197--7206. PMLR, 2020.

\bibitem{greedy-layer-reduction}
David Peer, Sebastian Stabinger, Stefan Engl, and Antonio
  Rodr{\'\i}guez-S{\'a}nchez.
\newblock Greedy-layer pruning: Speeding up transformer models for natural
  language processing.
\newblock {\em Pattern Recognition Letters}, 157:76--82, 2022.

\bibitem{qin2022bibert}
Haotong Qin, Yifu Ding, Mingyuan Zhang, Qinghua YAN, Aishan Liu, Qingqing Dang,
  Ziwei Liu, and Xianglong Liu.
\newblock Bi{BERT}: Accurate fully binarized {BERT}.
\newblock In {\em International Conference on Learning Representations}, 2022.

\bibitem{gpt-2}
Alec Radford, Jeff Wu, Rewon Child, David Luan, Dario Amodei, and Ilya
  Sutskever.
\newblock Language models are unsupervised multitask learners.
\newblock 2019.

\bibitem{T5}
Colin Raffel, Noam Shazeer, Adam Roberts, Katherine Lee, Sharan Narang, Michael
  Matena, Yanqi Zhou, Wei Li, and Peter~J. Liu.
\newblock Exploring the limits of transfer learning with a unified text-to-text
  transformer.
\newblock {\em CoRR}, abs/1910.10683, 2019.

\bibitem{rajpurkar2016squad}
Pranav Rajpurkar, Jian Zhang, Konstantin Lopyrev, and Percy Liang.
\newblock {SQuAD}: 100,000+ questions for machine comprehension of text.
\newblock {\em arXiv preprint arXiv:1606.05250}, 2016.

\bibitem{rastegari2016xnor}
Mohammad Rastegari, Vicente Ordonez, Joseph Redmon, and Ali Farhadi.
\newblock Xnor-net: Imagenet classification using binary convolutional neural
  networks.
\newblock In {\em European conference on computer vision}, pages 525--542.
  Springer, 2016.

\bibitem{romero2014fitnets}
Adriana Romero, Nicolas Ballas, Samira~Ebrahimi Kahou, Antoine Chassang, Carlo
  Gatta, and Yoshua Bengio.
\newblock Fitnets: Hints for thin deep nets.
\newblock {\em arXiv preprint arXiv:1412.6550}, 2014.

\bibitem{poorstable-fine-tune-bert-man-bert}
Hassan Sajjad, Fahim Dalvi, Nadir Durrani, and Preslav Nakov.
\newblock Poor man's {BERT:} smaller and faster transformer models.
\newblock {\em CoRR}, abs/2004.03844, 2020.

\bibitem{distill-bert}
Victor Sanh, Lysandre Debut, Julien Chaumond, and Thomas Wolf.
\newblock Distilbert, a distilled version of bert: smaller, faster, cheaper and
  lighter.
\newblock {\em CoRR}, abs/1910.01108, 2019.

\bibitem{sanh2020movement}
Victor Sanh, Thomas Wolf, and Alexander Rush.
\newblock Movement pruning: Adaptive sparsity by fine-tuning.
\newblock {\em Advances in Neural Information Processing Systems},
  33:20378--20389, 2020.

\bibitem{shen2020q}
Sheng Shen, Zhen Dong, Jiayu Ye, Linjian Ma, Zhewei Yao, Amir Gholami,
  Michael~W Mahoney, and Kurt Keutzer.
\newblock Q-bert: Hessian based ultra low precision quantization of bert.
\newblock In {\em Proceedings of the AAAI Conference on Artificial
  Intelligence}, 2020.

\bibitem{shomron2021post}
Gil Shomron, Freddy Gabbay, Samer Kurzum, and Uri Weiser.
\newblock Post-training sparsity-aware quantization.
\newblock {\em Advances in Neural Information Processing Systems}, 34, 2021.

\bibitem{mt-nlg}
Shaden Smith, Mostofa Patwary, Brandon Norick, Patrick LeGresley, Samyam
  Rajbhandari, Jared Casper, Zhun Liu, Shrimai Prabhumoye, George Zerveas,
  Vijay Korthikanti, et~al.
\newblock Using deepspeed and megatron to train megatron-turing nlg 530b, a
  large-scale generative language model.
\newblock {\em arXiv e-prints}, pages arXiv--2201, 2022.

\bibitem{socher2013recursive}
Richard Socher, Alex Perelygin, Jean Wu, Jason Chuang, Christopher~D Manning,
  Andrew~Y Ng, and Christopher Potts.
\newblock Recursive deep models for semantic compositionality over a sentiment
  treebank.
\newblock In {\em Proceedings of the 2013 conference on empirical methods in
  natural language processing}, pages 1631--1642, 2013.

\bibitem{patient-kd}
Siqi Sun, Yu~Cheng, Zhe Gan, and Jingjing Liu.
\newblock Patient knowledge distillation for {BERT} model compression.
\newblock In Kentaro Inui, Jing Jiang, Vincent Ng, and Xiaojun Wan, editors,
  {\em Proceedings of the 2019 Conference on Empirical Methods in Natural
  Language Processing and the 9th International Joint Conference on Natural
  Language Processing, {EMNLP-IJCNLP} 2019, Hong Kong, China, November 3-7,
  2019}, pages 4322--4331, 2019.

\bibitem{sun2020mobilebert}
Zhiqing Sun, Hongkun Yu, Xiaodan Song, Renjie Liu, Yiming Yang, and Denny Zhou.
\newblock Mobilebert: a compact task-agnostic bert for resource-limited
  devices.
\newblock In {\em Proceedings of the 58th Annual Meeting of the Association for
  Computational Linguistics}, pages 2158--2170, 2020.

\bibitem{how-to-trainbinary-nn}
Wei Tang, Gang Hua, and Liang Wang.
\newblock How to train a compact binary neural network with high accuracy?
\newblock In Satinder Singh and Shaul Markovitch, editors, {\em Proceedings of
  the Thirty-First {AAAI} Conference on Artificial Intelligence, February 4-9,
  2017, San Francisco, California, {USA}}, pages 2625--2631. {AAAI} Press,
  2017.

\bibitem{glue}
Alex Wang, Amanpreet Singh, Julian Michael, Felix Hill, Omer Levy, and
  Samuel~R. Bowman.
\newblock {GLUE:} {A} multi-task benchmark and analysis platform for natural
  language understanding.
\newblock In {\em 7th International Conference on Learning Representations},
  2019.

\bibitem{kd-review}
Lin Wang and Kuk{-}Jin Yoon.
\newblock Knowledge distillation and student-teacher learning for visual
  intelligence: {A} review and new outlooks.
\newblock {\em CoRR}, abs/2004.05937, 2020.

\bibitem{minilm-v2}
Wenhui Wang, Hangbo Bao, Shaohan Huang, Li~Dong, and Furu Wei.
\newblock Minilmv2: Multi-head self-attention relation distillation for
  compressing pretrained transformers.
\newblock {\em CoRR}, abs/2012.15828, 2020.

\bibitem{mini-lm}
Wenhui Wang, Furu Wei, Li~Dong, Hangbo Bao, Nan Yang, and Ming Zhou.
\newblock Minilm: Deep self-attention distillation for task-agnostic
  compression of pre-trained transformers.
\newblock In Hugo Larochelle, Marc'Aurelio Ranzato, Raia Hadsell,
  Maria{-}Florina Balcan, and Hsuan{-}Tien Lin, editors, {\em Advances in
  Neural Information Processing Systems 33: Annual Conference on Neural
  Information Processing Systems 2020, NeurIPS 2020, December 6-12, 2020,
  virtual}, 2020.

\bibitem{warstadt2018neural}
Alex Warstadt, Amanpreet Singh, and Samuel~R Bowman.
\newblock Neural network acceptability judgments.
\newblock {\em arXiv preprint arXiv:1805.12471}, 2018.

\bibitem{williams2017broad}
Adina Williams, Nikita Nangia, and Samuel~R Bowman.
\newblock A broad-coverage challenge corpus for sentence understanding through
  inference.
\newblock {\em arXiv preprint arXiv:1704.05426}, 2017.

\bibitem{yao2021mlpruning}
Zhewei Yao, Linjian Ma, Sheng Shen, Kurt Keutzer, and Michael~W Mahoney.
\newblock Mlpruning: A multilevel structured pruning framework for
  transformer-based models.
\newblock {\em arXiv preprint arXiv:2105.14636}, 2021.

\bibitem{yin2021autotinybert}
Yichun Yin, Cheng Chen, Lifeng Shang, Xin Jiang, Xiao Chen, and Qun Liu.
\newblock Autotinybert: Automatic hyper-parameter optimization for efficient
  pre-trained language models.
\newblock In {\em Proceedings of the 59th Annual Meeting of the Association for
  Computational Linguistics and the 11th International Joint Conference on
  Natural Language Processing (Volume 1: Long Papers)}, pages 5146--5157, 2021.

\bibitem{zadeh2020gobo}
Ali~Hadi Zadeh, Isak Edo, Omar~Mohamed Awad, and Andreas Moshovos.
\newblock Gobo: Quantizing attention-based nlp models for low latency and
  energy efficient inference.
\newblock In {\em 2020 53rd Annual IEEE/ACM International Symposium on
  Microarchitecture (MICRO)}, pages 811--824. IEEE, 2020.

\bibitem{zhang2020accelerating}
Minjia Zhang and Yuxiong He.
\newblock Accelerating training of transformer-based language models with
  progressive layer dropping.
\newblock {\em Advances in Neural Information Processing Systems},
  33:14011--14023, 2020.

\bibitem{zhang-etal-2020-ternarybert}
Wei Zhang, Lu~Hou, Yichun Yin, Lifeng Shang, Xiao Chen, Xin Jiang, and Qun Liu.
\newblock {T}ernary{BERT}: Distillation-aware ultra-low bit {BERT}.
\newblock In {\em Proceedings of the 2020 Conference on Empirical Methods in
  Natural Language Processing (EMNLP)}, pages 509--521, Online, November 2020.
  Association for Computational Linguistics.

\bibitem{extreme-small-vocab-kd}
Sanqiang Zhao, Raghav Gupta, Yang Song, and Denny Zhou.
\newblock Extremely small {BERT} models from mixed-vocabulary training.
\newblock In Paola Merlo, J{\"{o}}rg Tiedemann, and Reut Tsarfaty, editors,
  {\em Proceedings of the 16th Conference of the European Chapter of the
  Association for Computational Linguistics: Main Volume, {EACL} 2021, Online,
  April 19 - 23, 2021}, pages 2753--2759, 2021.

\bibitem{ad2}
Minjia Zhng, Niranjan~Uma Naresh, and Yuxiong He.
\newblock Adversarial data augmentation for task-specific knowledge
  distillation of pre-trained transformers.
\newblock {\em Association for the Advancement of Artificial Intelligence},
  2022.

\end{thebibliography}
}

\clearpage
\appendix
\section{Additional Related Work}\label{sec:KD-review}
KD has been extensively applied to computer vision and NLP tasks~\cite{kd-review} since its debut. On the NLP side, several variants of KD have been proposed to compress BERT~\cite{bert}, including how to define the knowledge that is supposed to be transferred from the teacher BERT model to the student variations. Examples of such knowledge definitions include output logits (e.g., DistilBERT~\cite{distill-bert}) and intermediate knowledge such as feature maps~\cite{patient-kd,kd-internal-amazon,extreme-small-vocab-kd} and self-attention maps~\cite{mini-lm,sun2020mobilebert} (we refer KD using these additional knowledge as deep knowledge distillation~\cite{mini-lm}). To mitigate the accuracy gap from reduced student model size, existing work has also explored applying knowledge distillation in the more expensive pre-training stage, which aims to provide a better initialization to the student for adapting to downstream tasks. As an example, MiniLM~\cite{mini-lm} and MobileBERT~\cite{sun2020mobilebert} advance the state-of-the-art by applying deep knowledge distillation and architecture change to pre-train a student model on the general-domain corpus, which then can be directly fine-tuned on downstream tasks with good accuracy. TinyBERT~\cite{jiao-etal-2020-tinybert} proposes to perform deep distillation in both the pre-training and fine-tuning stage and shows that these two stages of knowledge distillation are complementary to each other and can be combined to achieve state-of-the-art results on GLUE tasks. Our work differs from these methods in that we show how lightweight layer reduction without expensive pre-training distillation can be effectively combined with extreme quantization to achieve another order of magnitude reduction in model sizes for pre-trained Transformers.

\section{Additional Details on Methodology, Experimental Setup and Results}
\label{sec:method-app}

\subsection{Knowledge Distillation}
Knowledge Distillation (KD)~\cite{kd-hinton} has been playing the most  significant role in overcoming the performance degradation of model compression as the smaller models (i.e., student models) can absorb the rich knowledge of those uncompressed ones (i.e., teacher models) \cite{romero2014fitnets,lagunas2021block,sanh2020movement,gou2021knowledge}. In general, KD can be expressed as minimizing the difference ($\mathcal{L}_{kd}$) between outputs of a teacher model $T$ and a student model $S$.  As we are here studying transformer-based model and that the student and teacher models admit the same number (denoted as $h$) of attention heads, our setup focuses on a particular widely-used paradigm, layerwise KD \cite{jiao-etal-2020-tinybert,zhang-etal-2020-ternarybert,bai-etal-2021-binarybert}. That is, \emph{for each layer} of a student model $S$,  the loss in the knowledge transferred from teacher $T$ consist of three parts: 
\begin{itemize}
    \item[(1)]the fully-connected hidden states:  $\mathcal{L}_{\text{hidden}}=\textbf{MSE}(\mathbf{H}^S,\mathbf{H}^T )$, where $\mathbf{H}^S, \mathbf{H}^T\in \mathbb{R}^{l\times d_{h}}$;
    \item[(2)]the attention maps:  $\mathcal{L}_{\text{att}}=\frac{1}{h}\sum_{i=1}^h\textbf{MSE}(\mathbf{A}^S_i,\mathbf{A}^T_i)$, where  $\mathbf{A}^S_i, \mathbf{A}_i^T\in \mathbb{R}^{l\times l}$;
    \item[(3)]the prediction logits:  $\mathcal{L}_{\text{logit}}=\textbf{CE}(\mathbf{p}^S,\mathbf{p}^T)$ where  $\mathbf{p}^S, \mathbf{p}^T\in \mathbb{R}^{c}$; 
\end{itemize}
where $\mathbf{MSE}$ stands for the mean square error and $\mathbf{CE}$ is a cross entropy loss. Notably, for the first part in the above formulation, $\mathbf{H}^{S}$ ($\mathbf{H}^{T}$) corresponds to the output matrix of the student's (teacher's) hidden states; $l$ is the sequence length of the input (in our experiments, it is set to be $64$ or $128$); $d_h$ is the hidden dimension (768 for BERT$_{base}$). For the second part $\mathbf{A}_i^{S}$ ($\mathbf{A}_i^{T}$) is the attention matrix corresponds to the $i$-th heads (in our setting, $h=12$). In the final part, the dimension $c$ in logit outputs ($\mathbf{p}^S$ and $ \mathbf{p}^T$) is either to be $2$ or $3$ for GLUE tasks. For more details on how the weight matrices involved in the output matrices, please see \cite{jiao-etal-2020-tinybert}. 

 \begin{figure}
     \centering
\includegraphics[width=.8\textwidth]{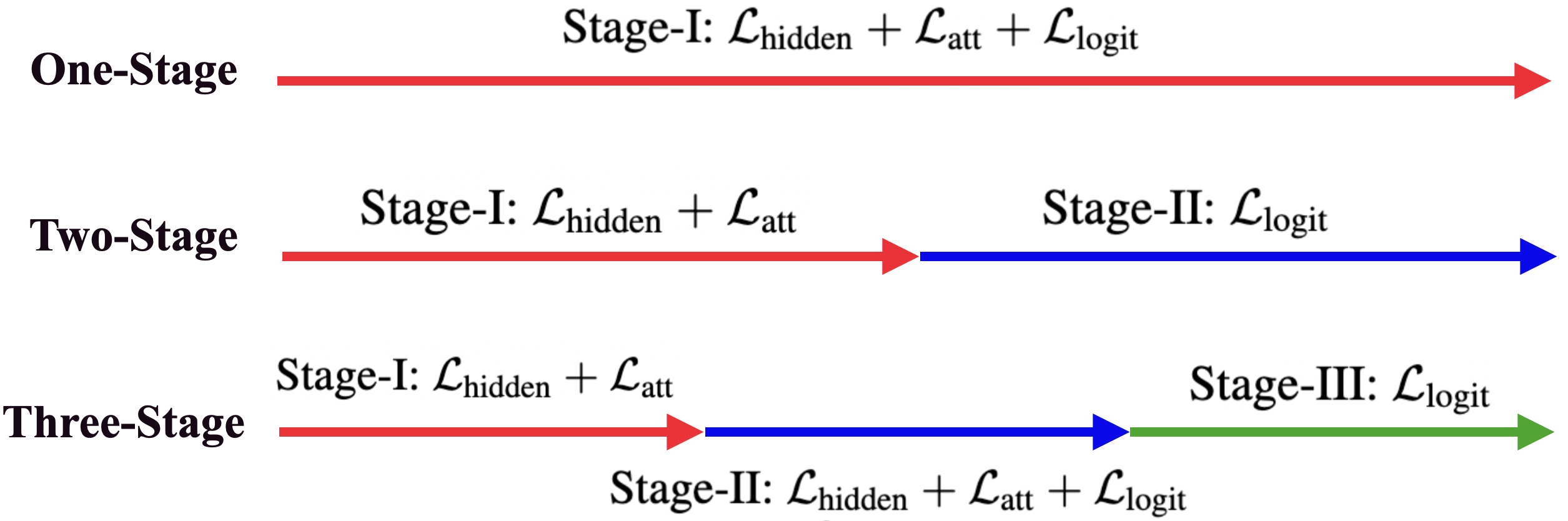}
\caption{\textbf{Three types of knowledge distillation.} \onestage KD (top red arrow line) involves all the outputs of hidden-states, attentions and logits from the beginning of the training to the end. \twostage KD (middle red and blue arrow line) separates hidden-states and attentions from the logits part. While \threestage KD (bottom red, blue and green arrow line) succeed \twostage one, it also adds a transition phase in the middle of the training.}
\label{fig:stage}
 \end{figure}

We have defined the three types of KD: \onestage, \twostage, and \threestage in main text. See \fref{fig:stage} for a visualization. Here we explain in more details. \textbf{One-Stage} KD means we naively minimize the sum of teacher-student differences on hidden-states, attentions and logits. In this setup, a single one-time learning rate schedule is used; in our case, it is a linear-decay schedule with warm-up steps 10\% of the total training time.  \textbf{Two-Stage} KD first minimizes the losses of hidden-states $\mathcal{L}_{\text{hidden}}$ and attentions $\mathcal{L}_{\text{att}}$, then followed by the loss of logits $\mathcal{L}_{\text{logit}}$. This type of KD is proposed in \cite{jiao-etal-2020-tinybert} and has been used in \cite{bai-etal-2021-binarybert} for extreme quantization. During the training, the (linear-decay) learning rate schedule  will be repeated from the first stage to the second one; particularly, in \cite{jiao-etal-2020-tinybert} they used a $\times 2.5$ peak learning rate for Stage-I than that for Stage-II. Finally, \textbf{Three-Stage} KD succeeds the properties of \onestage and \twostage. That is, after minimizing the losses of hidden-states $\mathcal{L}_{\text{hidden}}$ and attentions $\mathcal{L}_{\text{att}}$, we add a transition phase: $\mathcal{L}_{\text{hidden}}+\mathcal{L}_{\text{att}}+\mathcal{L}_{\text{logit}}$, instead of directly optimizing  $\mathcal{L}_{\text{logit}}$. The learning rate schedules will be correspondingly repeated in three times and the peak learning rate of Stage-I is $\times 2.5$ higher than that of Stage-II  and Stage-III.

\subsection{Experimental Setup}
 
Similar to \cite{jiao-etal-2020-tinybert,bai-etal-2021-binarybert}, we fine-tune on GLUE tasks, namely,
MRPC~\cite{dolan2005automatically},
STS-B~\cite{cer2017semeval},
SST-2~\cite{socher2013recursive}, QNLI~\cite{rajpurkar2016squad}, QQP~\cite{iyer2017first}, MNLI~\cite{williams2017broad}, CoLA~\cite{warstadt2018neural}, RTE~\cite{dagan2013recognizing}).   We recorded their validation performances over the training  and report the best validation value. The maximum sequence length is set to 64 for CoLA/SST-2, and 128 for the rest sequence pair tasks. See \tref{table:train-details} for the size of  each (augmented) dataset and training batch size. Each independent experiment is ran on a single A100 GPU with a fixed random seed 42. To understand how much the variance can bring with various random seed, we include a set of experiments by only varying the random seeds $111$ and $222$. The results are given in \tref{table:1-bit-bert-6-random-seeds} and \tref{table:2-bit-bert-6-random-seeds}. We overall find the standard deviation are within 0.1 on the GLUE score.

The 2-bit algorithm and distillation code are based on  the official implementation of \cite{zhang-etal-2020-ternarybert} \footnote{\small \url{https://github.com/huawei-noah/Pretrained-Language-Model/tree/master/TernaryBERT}} and the 1-bit algorithm is based only on the binary code\footnote{\small \url{https://github.com/huawei-noah/Pretrained-Language-Model/blob/master/BinaryBERT/transformer/utils\_quant.py\#L347}}  in  \cite{bai-etal-2021-binarybert} as explained in the main text. Finally, we comment on the reproduced result shown in Row 3 of \tref{table:pretrain}. We apply the \twostage where Stage-I distillation requires much longer epochs than Budget-C as ColA requires 50 epochs, MNLI/QQP/QNLI requires 10 epochs, and others 20 epochs. All are trained on augmented datasets; Note that 10-epoch augmented MNLI/QQP is much longer than 18/36-epoch standard ones. Stage-II KD is similar to Budget B but needs to search with three learning rates and two batch sizes.

\subsection{Exploration of LoRa}

 As we  see in \sref{sec:design} (i.e.,\tref{table:quantization-smaller-bert}), smaller models is much more brittle on extreme quantization than \bertbase, especially the 1-bit compression. We now investigate whether adding low-rank (LoRa) matrices to the quantize weights would help improve the situation. That is, we introduce two low-rank matrices $U\in\mathbb{R}^{d_{\text{in}}\times r}$ and  $V\in\mathbb{R}^{r\times d_{\text{out}}}$ for a quantized weight $\widetilde{W}\in\mathbb{R}^{d_{\text{in}}\times d_{\text{out}}}$ such that the weight we used for input is
 \begin{equation}
     W = \widetilde{W} +UV
 \end{equation}

 Here $U$ and $V$ are computed in full precision.
In our experiments for the layer-reduced model (Skip-Bert$_6$), we let  $r\in\{1,8\}$. We use \onestage and also include the scenarios for the three budgets and three learning rates \{5e-5,1e-4,5e-4\}. The full results for 1-bit and 2-bit  Skip-Bert$_6$ are presented in \tref{table:1-bit-bert-6-app} and \tref{table:2-bit-bert-6-app}. Correspondingly, we summarized the results in \tref{table:1-bit-bert-6} and \tref{table:2-bit-bert-6} for clarity.

Overall, we see that LoRa can boost performance when the training time is short. For instance, as shown in \tref{table:1-bit-bert-6} ( \tref{table:2-bit-bert-6}) under Budget-A, the average score for 1-bit (2-bit)  is 80.81 (82.14) without LoRa, compared with 81.09 (82.31) with LoRa. However, the benefit becomes marginal when training is long. Here, under Budget-B, the average score for 1-bit (2-bit)  is 81.37 (82.51) without LoRa, compared with 81.59 (82.41) with LoRa. 

To further verify the claim of "marginal or none" benefit using LoRa under a sufficient training budget, we include more experiments for five-/four-layer BERT (Skip-BERT$_5$ and Skip-BERT$_4$), and the results are shown in \tref{table:five-four-layer-quant}. The observation is similar to what we have found for Skip-BERT$_6$.

Interestingly, we find that in \tref{table:continue-1bit}, instead of adding these full-precision low-rank matrices,  there is a considerable improvement by continuing training with another Budget-C iteration based on the checkpoints obtained under Budget-C. Particularly, there is $0.09$ in GLUE score's improvement for 1-bit SkipBERT$_5$ (from  81.34 (Row 1) to 81.63 (Row 2)) and $0.28$ for 1-bit SkipBERT$_4$. On the other hand, if we trained a 2-bit SkipBERT$_5$/SkipBERT$_4$ first and then continue training them into 1-bit, the final performance  (see Row 5 and Row 11) is even worse than a direct 1-bit quantization (see Row 1 and Row 7). In addition to the results above, we also tried to see if continuing training by adding LoRa (rank=1) matrices would help or not. However, the results are negative.
\subsection{Exploration of Smaller Teacher}

As shown in previous works, such as \cite{minilm-v2, lagunas2021block}, there is a clear advantage to using larger teacher models. Here shown in \tref{table:teachers-q} under Budget-C, we apply 1-/2-bit quantization on Skip-BERT$_6$ and compare the teachers between \bertbase and the full-precision Skip-BERT$_6$.

We verify that a better teacher can boost the performance, particularly for 1-bit quantization on the smaller tasks such as CoLA, MRPC, RTE, and STS-B. However, when adding LoRa, the advantage of using better teacher diminishes as the gain on average GLUE score decreases from 0.31 (without LoRa) to 0.11 (with LoRa) on 1-/2-bit quantization.

\counterwithin{figure}{section}
\counterwithin{table}{section}
\section{Tables}\label{sec:table}
\begin{table}[ht]
\caption{Training details. DA is short for data augmentation.}\label{table:train-details}
\centering
\begin{adjustbox}{width=0.80\linewidth}
\begin{tabular}{lcccccccc}
\toprule
                 & CoLA   & MNLI      & MRPC    & QNLI    & QQP      & RTE     & SST-2   & STS-B\\\midrule
Dev data         & 1043   & 9815/9832 & 408     & 5463    & 40430    & 277     & 872     & 1500   \\
Train (noDA)     & 8551   & 392702    & 3668    & 104743  & 363846   & 2490    & 67349   & 5749   \\
Train (DA)       & 218844 & 392702    & 226123  & 4273773 & 363846   & 145865  & 1133834 & 327384 \\
DA/noDA          & 25.6   &  1        & 61.6    & 40.8    &  1       & 58.6    & 16.8    & 56.9   \\
Train batch-size & 16     & 32        & 32      & 32      & 32       & 32      & 32      & 32   \\
 \bottomrule
\end{tabular}
\end{adjustbox}
\end{table}

\begin{table}[ht]
\caption{1-bit quantization for BERT$_{\text{base}}$ with size $13.0$MB ($\times 32.0$ smaller than the fp32 version).}
\centering
\label{table:1-bit-bert-app}
\begin{adjustbox}{width=0.99\linewidth}
\centering
\begin{tabular}{lccccccccccccccc }
\toprule
Training & Stages    & learning    &  CoLA  & MNLI-m/-mm  &  MRPC      &  QNLI  &    QQP      & RTE    & SST-2 &   STS-B    & Avg. &Avg. w/o      \\
Cost &       &     rate &  Mcc     &   Acc/Acc   & F1/Acc     &  Acc   &  F1/Acc     & Acc    &  Acc  & Pear/Spea  &      &  CoLA   \\ 
\midrule
\multirow{12}{*}{Budget-A} 
           & \multirow{4}{*}{One-Stage}  &  2e-5  & 44.6  & 83.1/83.7  & 88.8/83.1  & 91.1  & 87.4/90.7  & 66.1  & 92.8  & 87.8/87.5  & 80.33 & 84.80 \\ 
                        &                & 1e-4   & 50.4  & 83.7/84.6  & 90.1/85.3  & 91.3  & 88.0/91.1  & 71.5  & 92.7  & 88.8/88.4  & 82.16 & 86.12 \\ 
                        &                & 5e-4   & 42.3  & 83.3/84.1  & 90.0/85.8  & 89.5  & 87.8/90.9  & 72.9  & 92.5  & 88.1/87.8  & 81.04 & 85.89 \\ \cdashline{3-13}
                     &           &  Best (above)  & 50.4  & 83.7/84.6  & 90.0/85.8  & 91.3  & 88.0/91.1  & 72.9  & 92.8  & 88.8/88.4  & 82.38 & 86.37 \\ \cline{2-13}
          & \multirow{4}{*}{Two-Stage}  &  2e-5  & 48.2  & 83.3/83.8  & 89.3/84.6  & 90.7  & 87.7/90.9  & 70.4  & 92.5  & 88.7/88.4  & 81.46 & 85.61 \\ 
                        &                & 1e-4   & 48.5  & 83.3/83.4  & 90.0/85.5  & 90.4  & 87.4/90.7  & 70.4  & 92.4  & 88.6/88.2  & 81.47 & 85.59 \\
\bottomrule
\end{tabular}
\end{adjustbox}
\end{table}

\begin{table}[ht]
\caption{2-bit quantization for BERT$_{\text{base}}$ with size $26.8$MB ($\times 16.0$ smaller than the fp32 version).}
\centering
\label{table:2-bit-bert-app}
\begin{adjustbox}{width=0.99\linewidth}
\centering
\begin{tabular}{lccccccccccccccc }
\toprule
Training & Stages    & learning    &  CoLA  & MNLI-m/-mm  &  MRPC      &  QNLI  &    QQP      & RTE    & SST-2 &   STS-B    & Avg. & Avg. w/o     \\
Cost &       &     rate &  Mcc     &   Acc/Acc   & F1/Acc     &  Acc   &  F1/Acc     & Acc    &  Acc  & Pear/Spea  &      &  CoLA   \\ 
\midrule
\multirow{12}{*}{Budget-A} 
           & \multirow{4}{*}{One-Stage}  &  2e-5  & 58.0  & 84.2/84.6  & 89.8/84.8  & 91.5  & 88.4/91.4  & 72.2  & 93.5  & 89.4/89.0  & 83.29 & 86.45 \\ 
                        &                & 1e-4   & 55.8  & 84.5/85.0  & 90.9/87.0  & 91.5  & 88.4/91.4  & 74.0  & 93.5  & 89.6/89.2  & 83.59 & 87.06 \\ 
                        &                & 5e-4   & 46.1  & 83.8/84.7  & 89.6/85.0  & 90.9  & 88.0/91.2  & 71.8  & 92.8  & 88.8/88.4  & 81.68 & 86.12 \\ \cdashline{3-13}
                     &           &  Best (above)  & 58.0  & 84.5/85.0  & 90.9/87.0  & 91.5  & 88.4/91.4  & 74.0  & 93.5  & 89.6/89.2  & 83.83 & 87.06 \\ \cline{2-13}
          & \multirow{4}{*}{Two-Stage}  &  2e-5  & 56.6  & 84.3/85.0  & 91.0/87.3  & 91.3  & 88.3/91.3  & 71.5  & 93.5  & 89.6/89.1  & 83.38 & 86.72 \\ 
                        &                & 1e-4   & 52.3  & 84.2/84.8  & 90.2/85.5  & 91.0  & 87.9/91.1  & 72.2  & 92.8  & 89.4/89.1  & 82.59 & 86.38 \\
\bottomrule
\end{tabular}
\end{adjustbox}
\end{table}

\begin{table}[ht]
\caption{Results without data augmentation. Details of \tref{table:da-1bit} for 1-bit quantization on \bertbase with three learning rates.}\label{table:da-1bit-app}
\begin{adjustbox}{width=0.99\linewidth}
\centering
\begin{tabular}{lccccccccccccccc }
\toprule
Cost& learning    &  CoLA  & MNLI-m/-mm  &  MRPC      &  QNLI  &    QQP      & RTE    & SST-2 &   STS-B    & Avg. &Avg. w/o     \\
  &  rate  & Mcc     &   Acc/Acc   & F1/Acc     &  Acc   &  F1/Acc     & Acc    &  Acc  & Pear/Spea  &    all  &  CoLA   \\   \midrule  
\multirow{2}{*}{Budget-A}  &  2e-5  & 47.3  & 83.1/83.7  & 87.2/80.1  & 90.8  & 87.4/90.7  & 66.8  & 92.2  & 85.5/85.3  & 80.02 & 84.11 \\ 
 \multirow{2}{*}{One-Stage}                        & 1e-4   & 48.9  & 83.7/84.6  & 89.0/83.6  & 91.0  & 88.0/91.1  & 71.5  & 92.5  & 87.6/87.5  & 81.61 & 85.70 \\ 
                         & 5e-4   & 38.1  & 83.3/84.1  & 88.4/83.6  & 90.1  & 87.8/90.9  & 69.7  & 92.0  & 85.0/85.0  & 79.64 & 84.84 \\ \cdashline{3-13}
                     &  Best (above)  & 48.9  & 83.7/84.6  & 89.0/83.6  & 91.0  & 88.0/91.1  & 71.5  & 92.5  & 87.6/87.5  & 81.61 & 85.70 \\
\midrule
\multirow{2}{*}{Budget-C} 
            &  2e-5  & 55.0  & 84.1/84.0  & 90.0/85.0  & 90.7  & 88.2/91.2  & 73.3  & 92.9  & 88.2/87.9  & 82.71 & 86.18 \\ 
 \multirow{2}{*}{One-Stage}                       & 1e-4   & 52.9  & 84.2/84.4  & 89.2/85.0  & 90.4  & 88.3/91.3  & 72.9  & 92.7  & 87.7/87.7  & 82.39 & 86.08 \\ 
                          & 5e-4   & 42.2  & 82.8/82.9  & 88.7/83.6  & 89.8  & 87.7/90.8  & 71.1  & 91.4  & 87.0/86.8  & 80.18 & 84.92 \\ \cdashline{3-13}
                     &  Best (above)  & 55.0  & 84.2/84.4  & 90.0/85.0  & 90.7  & 88.3/91.3  & 73.3  & 92.9  & 88.2/87.9  & 82.78 & 86.25 \\ 
 \bottomrule
\end{tabular}
\end{adjustbox}

\end{table}

\begin{table}[ht]
\caption{Details of \tref{table:pretrain}: Layer reduction with different learning rates (i.e., the 2nd column).} \label{table:pretrain-appendix}
\begin{adjustbox}{width=0.99\linewidth}
\centering
\begin{tabular}{lccccccccccccccc }
\toprule
Model      & Learning  &  CoLA  & MNLI-m/-mm  &  MRPC      &  QNLI  &    QQP      & RTE    & SST-2 &   STS-B    & Avg. &Avg. w/o     \\
(MB) &   rate  &  Mcc     &   Acc/Acc   & F1/Acc     &  Acc   &  F1/Acc     & Acc    &  Acc  & Pear/Spea  &      &  CoLA   \\   \midrule    
BERT-base             & n/a      & 59.7  &  84.9/85.6 & 90.6/86.3  & 92.1  & 88.6/91.5  & 72.2   & 93.2  & 90.1/89.6 & 83.95 & 86.98 \\\midrule
\multirow{3}{*}{TinyBERT$_{6}$}  & 5e-5 &  53.6  & 84.3/84.3  & 90.4/86.3  & 91.2  & 88.5/91.5  & 67.9  & 93.2  & 89.2/88.8  & 82.39 & 85.99 \\ 
                                &  1e-4 & 54.4  & 84.6/84.3  & 90.1/85.8  & 91.5  & 88.4/91.5  & 68.6  & 93.3  & 89.1/89.0  &  82.57 & 86.09 \\
                                &  5e-4 & 45.1  & 84.1/84.2  & 90.1/85.8  & 91.3  & 88.5/91.5  & 69.7  & 93.2  & 89.1/88.7  & 81.56 & 86.11 \\ \midrule   
\multirow{3}{*}{MiniLM$_{6}$}   & 5e-5  &  55.1  & 84.4/84.4  & 90.4/86.0  & 91.3  & 88.3/91.4  & 70.8  & 93.0  & 89.4/89.0  &  82.87 & 86.34 \\  
                                & 1e-4  &  55.4  & 84.2/84.5  & 89.5/85.3  & 91.3  & 88.4/91.5  & 70.8  & 93.2  & 89.4/89.0  & 82.84 & 86.28 \\
\bottomrule
\end{tabular}
\end{adjustbox}
\end{table}
\begin{table}[ht]
\caption{Selection for student networks from their  teacher networks (fine-tuned \bertbase). The following results follow the same hyperparamters and  the learning rate 5e-5.}\label{table:pretrain-student-slection}
\begin{adjustbox}{width=0.99\linewidth}
\centering
\begin{tabular}{llccccccccccccccc}
\toprule
Layer    &  CoLA  & MNLI-m/-mm  &  MRPC      &  QNLI  &    QQP      & RTE    & SST-2 &   STS-B    & Avg. &Avg. w/o     \\
Selection     &  Mcc     &   Acc/Acc   & F1/Acc     &  Acc   &  F1/Acc     & Acc    &  Acc  & Pear/Spea  &   all   & CoLA   \\   \midrule    
 BERT-{base} fp32             &  59.7  &  84.9/85.6   &  90.6/86.3  &  92.1  &  88.6/91.5   & 72.2   & 93.2  & 90.1/89.6 & 83.95 & 86.98\\\midrule
(1) Skip-6   &   56.9  & 84.5/84.9  & 89.7/84.8  & 91.7  & 88.5/91.5  & 71.8  & 93.5  & 89.8/89.4  & 83.27 & 86.56 \\ 
(2) Top-6     & 52.0 & 83.9/84.2 & 90.0/85.3 & 91.5 & 88.2/91.3  & 69.3 & 93.0 & 88.2/87.9 & 82.08 & 85.84 \\
(3) Bottom-6  & 51.2 & 83.2/83.6 & 89.5/84.6 & 90.1 & 87.8/90.9 & 67.9 & 92.7 & 89.5/89.1 & 81.50 & 85.29\\
 \bottomrule
\end{tabular}
\end{adjustbox}
\end{table}

 \begin{table}[ht]
\caption{Compare between different stages KD. All learning rate set to be 5e-5.}\label{table:stages-fp32}
\begin{adjustbox}{width=0.99\linewidth}
\centering
\begin{tabular}{lcccccccccccccccc }
\toprule
\# &  Model & \#-Stage   &  CoLA  & MNLI-m/-mm  &  MRPC      &  QNLI  &    QQP      & RTE    & SST-2 &   STS-B    & Avg. &Avg. w/o     \\
  &   (255.2MB) & Same Budget & Mcc     &   Acc/Acc   & F1/Acc     &  Acc   &  F1/Acc     & Acc    &  Acc  & Pear/Spea  &    all  &  CoLA   \\   \midrule    
1 &   \multirow{3}{*}{ TinyBERT$_{6}$}   & One  &  53.6  & 84.3/84.3  & 90.4/86.3  & 91.2  & 88.5/91.5  & 67.9  & 93.2  & 89.2/88.8  & 82.39 & 85.99 \\ 
2 &                                      & Two  &  55.3  & 84.2/84.2  & 89.9/85.3  & 91.1  & 88.4/91.4  & 68.6  & 92.9  & 89.2/88.9  & 82.47 & 85.86 \\ 
3 &                                      & Three  &  54.2  & 84.2/84.4  & 89.9/85.0  & 91.3  & 88.5/91.5  & 70.8  & 93.1  & 89.2/88.8  & 82.63 & 86.19 \\ \midrule 
4 & \multirow{3}{*}{MiniLM$_6$-v2}  & One            &  55.1  & 84.4/84.4  & 90.4/86.0  & 91.3  & 88.3/91.4  & 70.8  & 93.0  & 89.4/89.0  & 82.87 & 86.34 \\  
5 &                            & Two            &  53.8  & 84.4/84.5  & 90.7/87.3  & 91.2  & 88.3/91.3  & 67.9  & 93.1  & 89.7/89.3  & 82.58 & 86.18 \\ 
6 &                            & Three            &  54.0  & 84.4/84.6  & 89.1/85.3  & 91.3  & 88.5/91.5  & 70.4  & 93.0  & 89.6/89.2  & 82.68 & 86.26 \\  \midrule  
7 &  \multirow{3}{*}{Skip-BERT$_{6}$}  & One      & 56.9  & 84.5/84.9  & 89.7/84.8  & 91.7  & 88.5/91.5  & 71.8  & 93.5  & 89.8/89.4  & 83.27 & 86.56 \\ 
8 &                                  & Two      & 56.0  & 84.1/84.2  & 89.7/84.8  & 91.5  & 88.2/91.3  & 71.8  & 93.3  & 89.4/89.1  & 82.93 & 86.30 \\  
9 &                                  & Three      & 55.3  & 84.4/84.6  & 89.7/84.8  & 91.7  & 88.5/91.5  & 72.2  & 93.6  & 89.6/89.2  & 83.08 & 86.55 \\ 
 \bottomrule
\end{tabular}
\end{adjustbox}
\end{table}

\begin{table}[ht]
\caption{1-/2-bit quantization of smaller model.}\label{table:quantization-smaller-bert-app}
\begin{adjustbox}{width=0.99\linewidth}
\centering
\begin{tabular}{lccccccccccccccc }
\toprule
 & Model &     & size    &  CoLA  & MNLI-m/-mm  &  MRPC      &  QNLI  &    QQP      & RTE    & SST-2 &   STS-B    & Avg. & Acc.   \\
\#& \#-layer &bit   &   (MB) & Mcc     &   Acc/Acc   & F1/Acc     &  Acc   &  F1/Acc     & Acc    &  Acc  & Pear/Spea  &   all   & drop \\   \midrule    
1 & \multirow{2}{*}{TinyBERT$_{6}$\cite{zhang-etal-2020-ternarybert}} & 32 & 255.2 ($\times 1.6$) & 54.1  &  84.5/84.5   &  91.0/87.3  &  91.1  &  88.0/91.1   & 71.8   & 93.0  & 89.8/89.6 & 83.02 & --\\
2 &                                                                   & 2  & 16.0 ($\times 26.2$)  & 53.0  & 83.4/83.8  & 91.5/88.0  & 89.9  & 87.2/90.5  & 71.8  & 93.0  & 86.9/86.5  & 82.26 & -0.76 \\    \midrule
3&\multirow{2}{*}{\OURS-BERT$_6$} & 32 & 255.2 ($\times 1.6$)   &  56.9  &  84.4/84.8   &  90.1/85.5  & 91.3  &  88.4/91.4   & 72.2   & 93.2  & 90.3/90.0   & 83.33 & -- \\\cdashline {3-14}
4& \multirow{2}{*}{SIX}                   & {2}    & 16.0 ($\times 26.2$)  & 53.8  & 83.6/84.2  & 90.5/86.3  & 90.6  & 88.2/91.3  & 73.6  & 93.6  & 89.0/88.7  & 82.89 & -0.44 \\  
5&                   &  {1}    & 8.0 ($\times 52.3$)    & 52.3  & 83.4/83.8  & 90.0/85.3  & 89.4  & 87.9/91.1  & 68.6  & 93.1  & 88.4/88.0  & 81.71 & -1.51 \\   \midrule
6& \multirow{2}{*}{\OURS-BERT$_5$}& 32 & 228.2 ($\times 1.8$)   & 56.7  & 84.2/84.8  & 90.1/85.5  & 91.4  & 88.3/91.4  & 72.2  & 93.2  & 89.2/88.8  & 83.18 & --  \\
\bottomrule
\end{tabular}
\end{adjustbox}
\end{table}

\begin{table}[ht]
\caption{\textbf{1-bit} quantization for Skip-BERT$_{6}$.}
\centering
\label{table:1-bit-bert-6}
\begin{adjustbox}{width=0.99\linewidth}
\centering
\begin{tabular}{llcccccccccccccc }
\toprule
\# & Cost    & LoRa    &  CoLA  & MNLI-m/-mm  &  MRPC      &  QNLI  &    QQP      & RTE    & SST-2 &   STS-B    & Avg. &Avg. w/o      \\
&       &  (Y/N) & Mcc     &   Acc/Acc   & F1/Acc     &  Acc   &  F1/Acc     & Acc    &  Acc  & Pear/Spea  &      &  CoLA   \\ \midrule
1 &\multirow{3}{*}{Budget-A} 
             &      N   & 47.9  & 82.9/83.2  & 90.5/86.3  & 89.7  & 87.8/90.9  & 65.7  & 92.7  & 88.0/87.6  & 80.81 & 84.92 \\ 
2 &          &      Y (DIM=1)   & 48.1  & 82.9/83.2  & 89.4/84.3  & 89.6  & 87.6/90.8  & 65.3  & 92.4  & 88.1/87.7  & 80.52 & 84.58 \\ 
3 &          &      Y (DIM=8)   & 49.7  & 83.4/83.6  & 89.8/85.0  & 89.9  & 87.5/90.8  & 66.8  & 92.8  & 87.8/87.5  & 81.09 & 85.01 \\ 
\midrule
4 &\multirow{3}{*}{Budget-B} 
             &      N   & 50.3  & 83.2/83.7  & 90.9/86.8  & 89.3  & 87.9/91.0  & 67.1  & 92.7  & 88.2/88.0  & 81.37 & 85.25 \\ 
5 &          &      Y (DIM=1)   & 49.9  & 83.3/83.8  & 90.5/86.0  & 89.3  & 88.0/91.1  & 67.5  & 92.7  & 88.3/87.9  & 81.32 & 85.25 \\ 
6 &          &      Y (DIM=8)   & 50.9  & 83.2/83.7  & 90.3/85.8  & 89.4  & 88.1/91.2  & 69.0  & 92.8  & 88.3/88.0  & 81.59 & 85.42 \\
\bottomrule
\end{tabular}
\end{adjustbox}
\end{table}

\begin{table}[ht]
\caption{\textbf{2-bit} quantization for Skip-BERT$_{6}$.}
\centering
\label{table:2-bit-bert-6}
\begin{adjustbox}{width=0.99\linewidth}
\centering
\begin{tabular}{llcccccccccccccc }
\toprule
\# & Cost    & LoRa    &  CoLA  & MNLI-m/-mm  &  MRPC      &  QNLI  &    QQP      & RTE    & SST-2 &   STS-B    & Avg. &Avg. w/o      \\
&      & (Y/N) & Mcc     &   Acc/Acc   & F1/Acc     &  Acc   &  F1/Acc     & Acc    &  Acc  & Pear/Spea  &      &  CoLA   \\ \midrule
1 &\multirow{3}{*}{Budget-A} 
             &      N   & 51.8  & 83.8/83.7  & 90.2/85.8  & 90.6  & 88.2/91.3  & 70.4  & 93.2  & 88.7/88.3  & 82.14 & 85.94 \\ 
2 &          &      Y (DIM=1)   & 51.4  & 83.7/84.2  & 90.3/85.8  & 90.5  & 88.2/91.3  & 70.8  & 93.5  & 88.9/88.5  & 82.23 & 86.09 \\ 
3 &          &      Y (DIM=8)   & 51.6  & 83.6/84.2  & 90.6/86.3  & 90.7  & 88.2/91.3  & 71.1  & 93.1  & 88.9/88.5  & 82.31 & 86.15 \\ 
\midrule
4 &\multirow{3}{*}{Budget-B} 
             &      N   & 52.7  & 83.8/84.5  & 90.0/85.3  & 90.3  & 88.4/91.4  & 72.6  & 93.0  & 89.0/88.6  & 82.51 & 86.24 \\ 
5 &          &      Y (DIM=1)   & 53.6  & 83.6/83.9  & 90.5/86.0  & 90.2  & 88.4/91.4  & 71.1  & 93.2  & 88.9/88.5  & 82.43 & 86.04 \\ 
6 &          &      Y (DIM=8)   & 53.3  & 83.9/84.4  & 90.3/85.8  & 90.2  & 88.3/91.3  & 70.8  & 93.0  & 89.0/88.7  & 82.41 & 86.05 \\ 
\bottomrule
\end{tabular}
\end{adjustbox}
\end{table}

\begin{table}[ht]
\caption{\textbf{1-bit} quantization for Skip-BERT$_{6}$.} 
\centering
\label{table:1-bit-bert-6-app}
\begin{adjustbox}{width=0.99\linewidth}
\centering
\begin{tabular}{lccccccccccccccc }
\toprule
Training & LoRa    & learning    &  CoLA  & MNLI-m/-mm  &  MRPC      &  QNLI  &    QQP      & RTE    & SST-2 &   STS-B    & Avg. &Avg. w/o     \\
Cost &       &     rate &  Mcc     &   Acc/Acc   & F1/Acc     &  Acc   &  F1/Acc     & Acc    &  Acc  & Pear/Spea  &      &  CoLA   \\ 
\midrule
\multirow{12}{*}{Budget-A} 
           & \multirow{4}{*}{N}  &  5e-5  & 45.6  & 82.6/83.0  & 89.3/84.3  & 89.7  & 87.5/90.8  & 63.5  & 92.4  & 87.6/87.3  & 79.94 & 84.24 \\ 
                        &                & 1e-4   & 47.9  & 82.9/83.2  & 89.4/84.6  & 89.4  & 87.8/90.9  & 64.3  & 92.4  & 88.0/87.6  & 80.40 & 84.46 \\ 
                        &                & 5e-4   & 40.6  & 81.9/82.6  & 90.5/86.3  & 88.4  & 87.6/90.8  & 65.7  & 92.7  & 86.9/86.7  & 79.54 & 84.41 \\ \cdashline{3-13}
                     &           &  Best (above)  & 47.9  & 82.9/83.2  & 90.5/86.3  & 89.7  & 87.8/90.9  & 65.7  & 92.7  & 88.0/87.6  & 80.81 & 84.92 \\ \cline{2-13}
                        & \multirow{4}{*}{Y (DIM=1)}  &  5e-5  & 46.9  & 82.9/83.2  & 88.9/83.6  & 89.6  & 87.6/90.8  & 63.2  & 92.4  & 87.7/87.4  & 80.03 & 84.18 \\ 
                        &                & 1e-4   & 48.1  & 82.6/83.5  & 89.4/84.3  & 89.2  & 87.6/90.8  & 65.3  & 92.3  & 88.1/87.7  & 80.47 & 84.51 \\
\bottomrule
\end{tabular}
\end{adjustbox}
\end{table}

\begin{table}[ht]
\caption{\textbf{2-bit} quantization for Skip-BERT$_{6}$ with various learning rates.}
\centering
\label{table:2-bit-bert-6-app}
\begin{adjustbox}{width=0.99\linewidth}
\centering
\begin{tabular}{lccccccccccccccc }
\toprule
Training & LoRa    & learning    &  CoLA  & MNLI-m/-mm  &  MRPC      &  QNLI  &    QQP      & RTE    & SST-2 &   STS-B    & Avg. &Avg. w/o      \\
Cost &       &     rate &  Mcc     &   Acc/Acc   & F1/Acc     &  Acc   &  F1/Acc     & Acc    &  Acc  & Pear/Spea  &      &  CoLA   \\ 
\midrule
\multirow{12}{*}{Budget-A} 
           & \multirow{4}{*}{N}  &  5e-5  & 50.8  & 83.8/83.7  & 90.0/85.3  & 90.4  & 88.2/91.3  & 70.4  & 93.2  & 88.6/88.2  & 81.94 & 85.84 \\ 
                        &                & 1e-4   & 51.8  & 83.6/84.2  & 90.2/85.8  & 90.6  & 88.2/91.3  & 70.4  & 92.9  & 88.7/88.3  & 82.14 & 85.94 \\ 
                        &                & 5e-4   & 41.4  & 83.2/83.5  & 90.1/85.5  & 89.8  & 87.9/91.0  & 68.2  & 92.5  & 87.8/87.4  & 80.32 & 85.19 \\ \cdashline{3-13}
                     &           &  Best (above)  & 51.8  & 83.8/83.7  & 90.2/85.8  & 90.6  & 88.2/91.3  & 70.4  & 93.2  & 88.7/88.3  & 82.14 & 85.94 \\ \cline{2-13}
                        & \multirow{4}{*}{Y (DIM=1)}  &  5e-5  & 51.4  & 83.7/84.2  & 90.3/85.8  & 90.5  & 88.2/91.3  & 70.8  & 93.5  & 88.8/88.4  & 82.22 & 86.07 \\ 
                        &                & 1e-4   & 51.2  & 83.7/84.0  & 90.3/85.8  & 90.4  & 88.1/91.2  & 70.8  & 92.9  & 88.9/88.5  & 82.10 & 85.96 \\
\bottomrule
\end{tabular}
\end{adjustbox}
\end{table}
\begin{table}[ht]
\caption{LoRa and quantization of five-/four-layer model.}
\label{table:five-four-layer-quant}
\begin{adjustbox}{width=0.99\linewidth}
\centering
\begin{tabular}{lcccccccccccccccc }
\toprule
\#- & \#- & loRa   & size    &  CoLA  & MNLI-m/-mm  &  MRPC      &  QNLI  &    QQP      & RTE    & SST-2 &   STS-B    & Avg. &Avg. w/o    \\
 layer  & bit  & Y/N    & (MB) & Mcc     &   Acc/Acc   & F1/Acc     &  Acc   &  F1/Acc     & Acc    &  Acc  & Pear/Spea  &   all   &  CoLA   \\   \midrule    
\multirow{5}{*}{Five}& 32 & & 228.2 ($\times 1.8$)   & 56.7  & 84.2/84.8  & 90.1/85.5  & 91.4  & 88.3/91.4  & 72.2  & 93.2  & 89.2/88.8  & 83.18 & 86.49   \\ \cdashline {2-14}
                    & \multirow{2}{*}{1} & N & 7.1 ($\times 58.5$) &  52.2  & 82.9/83.2  & 89.9/85.0  & 88.5  & 87.6/90.8  & 69.3  & 92.9  & 87.3/87.0  & 81.34 & 84.99 \\  
                    &                    & Y &7.3 ($\times 57.4$)   & 51.8  & 81.8/81.9  & 90.2/85.5  & 88.6  & 87.1/90.3  & 68.6  & 92.8  & 87.5/87.2  & 80.98 & 84.62 \\
\bottomrule
\end{tabular}
\end{adjustbox}
\end{table}

\begin{table}
\caption{More quantization of five-/four-layer model.}\label{table:continue-1bit}
\begin{adjustbox}{width=0.99\linewidth}
\centering
\begin{tabular}{llcccccccccccccc }
\toprule
\# & Model    &  size    &  CoLA  & MNLI-m/-mm  &  MRPC      &  QNLI  &    QQP      & RTE    & SST-2 &   STS-B    & Avg. &Avg. w/o     \\
&    &      &  Mcc     &   Acc/Acc   & F1/Acc     &  Acc   &  F1/Acc     & Acc    &  Acc  & Pear/Spea  &      &  CoLA   \\   \midrule    
1 &  1-bit  SkipBERT$_5$   & 7.1 ($\times 58.5$) &  52.2  & 82.9/83.2  & 89.9/85.0  & 88.5  & 87.6/90.8  & 69.3  & 92.9  & 87.3/87.0  & 81.34 & 84.99 \\  
\multicolumn{6}{l}{Continue with above  using another one-stage}\\
2 &  1-bit      & 7.1 ($\times 58.5$) & 52.8  & 83.0/83.4  & 89.8/84.8  & 89.6  & 87.8/91.0  & 69.3  & 93.2  & 87.6/87.3  & 81.63 & 85.24 \\  
3 & 1-bit+LoRa1      & 7.3 ($\times 57.4$)  & 53.0  & 83.0/83.5  & 89.5/84.6  & 89.6  & 87.9/91.0  & 69.3  & 93.2  & 87.6/87.3  & 81.64 & 85.23 \\ 
\cdashline {1-14}
4 &  2-bit SkipBERT$_5$      & 14.2 ($\times 29.3$) & 53.9  & 83.3/84.1  & 90.4/86.0  & 90.4  & 88.2/91.2  & 71.8  & 93.0  & 88.4/88.0  & 82.46 & 86.02 \\ 
\multicolumn{6}{l}{Continue with above using another one-stage}\\
\bottomrule
\end{tabular}
\end{adjustbox}
\end{table}

\begin{table}
\caption{Compare between teachers: a twelve-layer \bertbase vs a six-layer SkipBERT$_{6}$.}
\label{table:teachers-q}
\begin{adjustbox}{width=0.99\linewidth}
\centering
\begin{tabular}{lccccccccccccccccc}
\toprule
\#-    &  LoRa (Size) & Teacher & CoLA  & MNLI-m/-mm  &  MRPC      &  QNLI  &    QQP      & RTE    & SST-2 &   STS-B    & Avg. &Avg. w/o     \\
bit    &   Y/N (MB)      &    & Mcc     &   Acc/Acc   & F1/Acc     &  Acc   &  F1/Acc     & Acc    &  Acc  & Pear/Spea  &      &  CoLA   \\   \midrule    
\multirow{6}{*}{1-bit} & \multirow{2}{*}{N (8.0MB)}  & \bertbase  & 52.3  & 83.4/83.8  & 90.0/85.3  & 89.4  & 87.9/91.1  & 68.6  & 93.1  & 88.4/88.0  & 81.71 & 85.39 \\  
                       &                             & Skip-BERT$_6$    & 52.0  & 83.2/83.5  & 89.7/85.0  & 89.3  & 87.7/91.0  & 67.5  & 93.2  & 87.9/87.6  & 81.40 & 85.08 \\  \cdashline{2-14}
                       & \multicolumn{2}{c}{Difference}                        & +0.3  & +0.2/+0.3  & +0.3/+0.3  & +0.1  & +0.2/+0.1  & +1.1  & -0.1  & +0.5/+0.4  & +0.31 & +0.31 \\  \cline{2-14}
                       &  \multirow{2}{*}{Y (8.1MB)} & \bertbase  & 52.1  & 83.3/84.0  & 89.9/85.5  & 89.5  & 88.0/91.1  & 68.2  & 93.0  & 88.5/88.1  & 81.69 & 85.39    \\  
                       &                             & Skip-BERT$_6$    & 52.4  & 83.2/83.5  & 90.4/85.8  & 89.1  & 87.8/91.0  & 68.2  & 93.1  & 87.9/87.6  & 81.58 & 85.22 \\  \cdashline{2-14}
                       & \multicolumn{2}{c}{Difference}               & -0.3  & +0.1/+0.5  & -0.5/-0.3  & +0.1  & +0.2/+0.1  & 0.0   & -0.1  & +0.6/+0.5  & +0.11 & +0.17 \\
\bottomrule
\end{tabular}
\end{adjustbox}
\end{table}

\begin{table}[ht]
\caption{\textbf{1-bit} quantization with various random seeds (learning rate here all set to be 2e-5).} 
\centering
\label{table:1-bit-bert-6-random-seeds}
\begin{adjustbox}{width=0.99\linewidth}
\centering
\begin{tabular}{lccccccccccccccc }
\toprule
Model & KD    & Random    &  CoLA  & MNLI-m/-mm  &  MRPC      &  QNLI  &    QQP      & RTE    & SST-2 &   STS-B    & Avg.     \\
Cost&  \#-Stage     &    seed &  Mcc     &   Acc/Acc   & F1/Acc     &  Acc   &  F1/Acc     & Acc    &  Acc  & Pear/Spea  &   \\ 
\midrule
                      
\multirow{11}{*}{TinyBERT$_6$} & \multicolumn{2}{c}{fp32} & 54.2 & 84.25/84.38 & 89.85/85.05 & 91.3 & 88.49/91.45 & 70.8 & 93.1 & 89.18/88.79 & 82.64 \\\cline{2-13}
\multirow{11}{*}{Budget-B}   &  \multirow{5}{*}{One}     & 42   &50.5 & 82.72/82.95 & 89.08/84.07 & 89.3 & 87.72/90.82 & 65.0 & 92.7 & 87.93/87.64 & 80.66 \\
                           &    & 111   & 49.6 & 82.75/83.15 & 89.04/84.07 & 89.4 & 87.8/90.97  & 65.0 & 92.4 & 87.95/87.75 & 80.59 \\
 &    & 222   & 48.4 & 82.72/83.08 & 89.11/84.07 & 89.4 & 87.61/90.84 & 64.6 & 92.7 & 88.14/87.75 & 80.43 \\\cdashline{3-13}
 &   \multicolumn{2}{r}{Ave. above}  & 49.5 & 82.73/83.06 & 89.08/84.07 & 89.4 & 87.71/90.88 & 64.9 & 92.6 & 88.01/87.71 & 80.56 \\
 &    \multicolumn{2}{r}{std  above} &  0.9  & 0.01/0.08   & 0.03/0      & 0.0  & 0.08/0.07   & 0.2  & 0.1  & 0.09/0.05   & 0.09          \\\cline{2-13}             
  & \multirow{5}{*}{Two}  &  42 &48.8 & 82.58/82.81 & 88.74/83.58 & 89.2 & 87.75/90.87 & 65.0 & 92.4 & 88.2/87.82  & 80.37 \\
 &                &111   &48.9 & 82.84/83.31 & 88.93/83.82 & 89.0 & 87.72/90.87 & 64.3 & 92.7 & 88.4/88.06  & 80.45 \\
&                & 222  & 48.6 & 82.72/83.27 & 89.56/84.8  & 88.9 & 87.68/90.89 & 65.7 & 92.6 & 88.2/87.87  & 80.62 \\\cdashline{3-13}
  &   \multicolumn{2}{r}{Ave. above}  & 48.8 & 82.71/83.13 & 89.08/84.07 & 89.0 & 87.72/90.88 & 65.0 & 92.6 & 88.27/87.92 & 80.48 \\
 &    \multicolumn{2}{r}{std  above}  &0.1  & 0.11/0.23   & 0.35/0.53   & 0.1  & 0.03/0.01   & 0.6  & 0.1  & 0.09/0.1    & 0.10 \\
 \midrule
\multirow{6}{*}{MiniLM$_6$} 
        &  \multicolumn{2}{c}{fp32}   & 54.0 & 84.42/84.58 & 89.13/85.29 & 91.3 & 88.53/91.46 & 70.4 & 93.0 & 89.56/89.19 & 82.67\\\cline{2-13}
\multirow{6}{*}{Budget-B}    & \multirow{5}{*}{Three} &    42 & 48.2 & 83.03/83.24 & 88.47/83.33 & 89.2 & 87.85/90.87 & 64.3 & 92.6 & 88.37/88.06 & 80.34 \\
 &       & 111   &  47.7 & 83.2/83.42  & 88.56/82.84 & 89.4 & 87.76/90.9  & 65.3 & 92.6 & 88.48/88.17 & 80.43 \\
                          &    & 222   &  47.2 & 83.36/83.44 & 89.04/83.82 & 89.4 & 87.88/90.97 & 64.6 & 92.6 & 88.57/88.23 & 80.43 \\ \cdashline{3-13}
 &   \multicolumn{2}{r}{Ave. above}  &  47.7 & 83.2/83.37  & 88.69/83.33 & 89.3 & 87.83/90.91 & 64.7 & 92.6 & 88.47/88.15 & 80.40 \\
 &    \multicolumn{2}{r}{std  above} & 0.4  & 0.13/0.09   & 0.25/0.4    & 0.1  & 0.05/0.04   & 0.5  & 0.0  & 0.08/0.07   & 0.04 \\\midrule
\multirow{6}{*}{SkipBERT$_6$}  &  \multicolumn{2}{c}{fp32}   &  56.1 & 84.37/84.53 & 90.52/86.03 & 91.6 & 88.34/91.42 & 73.7 & 93.5 & 89.3/88.92 & 83.39\\\cline{2-13}
 \multirow{6}{*}{Budget-B} & \multirow{4}{*}{Three}  & 42  & 51.1 & 82.79/83.59 & 89.95/85.05 & 89.4 & 88.01/91.07 & 67.5 & 92.9 & 88.09/87.73 & 81.28 \\
  &                  & 111 & 50.7 & 82.96/83.2  & 90.02/85.54 & 89.4 & 87.96/91.04 & 67.9 & 92.9 & 88.35/87.99 & 81.32 \\
 &                  &222&  51.8 & 83.02/83.58 & 90.54/86.27 & 89.4 & 87.91/91    & 67.2 & 92.6 & 88.23/87.98 & 81.43 \\ \cdashline{3-13}
 &   \multicolumn{2}{r}{Ave. above}  &  51.2 & 82.92/83.46 & 90.17/85.62 & 89.4 & 87.96/91.04 & 67.5 & 92.8 & 88.22/87.9  & 81.35 \\
 &    \multicolumn{2}{r}{std  above} & 0.4  & 0.1/0.18    & 0.26/0.5    & 0.0  & 0.04/0.03   & 0.3  & 0.2  & 0.11/0.12   & 0.06    &  \\                   
\bottomrule
\end{tabular}
\end{adjustbox}
\end{table}

\begin{table}[ht]
\caption{\textbf{2-bit} quantization with various random seeds (learning rate here all set to be 2e-5).} 
\centering
\label{table:2-bit-bert-6-random-seeds}
\begin{adjustbox}{width=0.99\linewidth}
\centering
\begin{tabular}{lccccccccccccccc }
\toprule
Model & KD    & Random    &  CoLA  & MNLI-m/-mm  &  MRPC      &  QNLI  &    QQP      & RTE    & SST-2 &   STS-B    & Avg.     \\
Cost&  \#-Stage     &    seed &  Mcc     &   Acc/Acc   & F1/Acc     &  Acc   &  F1/Acc     & Acc    &  Acc  & Pear/Spea  &   \\ 
\midrule
\multirow{4}{*}{SkipBERT$_6$} 
 & \multirow{4}{*}{Three}  & 42  &  53.2 & 83.67/83.79 & 91.25/87.25 & 90.4 & 88.42/91.4  & 72.2 & 93.6 & 88.77/88.36 & 82.70 \\
\multirow{4}{*}{Budget-B}  &                  & 111 &  53.3 & 83.63/83.59 & 90.91/86.76 & 90.4 & 88.32/91.38 & 72.2 & 93.5 & 88.65/88.32 & 82.60 \\
 &                  & 222 & 54.1 & 83.65/84.27 & 90.82/86.52 & 90.6 & 88.29/91.4  & 71.5 & 93.6 & 88.69/88.38 & 82.70 \\\cdashline{3-13}
 &   \multicolumn{2}{r}{Ave. above}  & 53.6 & 83.65/83.88 & 90.99/86.84 & 90.4 & 88.34/91.39 & 72.0 & 93.5 & 88.7/88.35  & 82.66 \\ 
 &    \multicolumn{2}{r}{std  above} & 0.4  & 0.02/0.29   & 0.19/0.3    & 0.1  & 0.06/0.01   & 0.3  & 0.1  & 0.05/0.02   & 0.05 \\                   
\bottomrule
\end{tabular}
\end{adjustbox}
\end{table}

\end{document}